\documentclass{article}

\usepackage[main, final]{neurips_2026}
\usepackage{mathtools}
\usepackage{dsfont}
\usepackage{algorithm}
\usepackage{algorithmic}
\usepackage{natbib}
\usepackage[utf8]{inputenc} 
\usepackage[T1]{fontenc}    
\usepackage{url}            
\usepackage{booktabs}       
\usepackage{amsfonts}       
\usepackage{nicefrac}       
\usepackage{microtype}      
\usepackage{xcolor}         
\usepackage{multirow}
\usepackage{graphicx}
\usepackage{wrapfig}
\usepackage{amsmath}
\usepackage{tcolorbox}
\tcbuselibrary{breakable}
\usepackage{tabularx}
\usepackage{booktabs}
\usepackage{array}
\usepackage{bm}
\usepackage{caption}
\usepackage{fontawesome5}
\usepackage[table]{xcolor}

\definecolor{GroupGray}{RGB}{242,242,242}

\newcommand{\ie}{\textit{i.e.}}
\newcommand{\eg}{\textit{e.g.}}

\definecolor{cvprblue}{rgb}{0.21,0.49,0.74}
\usepackage[pagebackref,breaklinks,colorlinks,citecolor=cvprblue]{hyperref}

\definecolor{RedOrange}{rgb}{1.0, 0.27, 0.0}
\newcommand{\ourup}[1]{$_{\color{RedOrange}\uparrow \textbf{#1}}$}
\newcommand{\ourdown}[1]{$_{\color{blue!25!gray} \downarrow #1}$}

\title{VerifyMAS: Hypothesis Verification for Failure Attribution in LLM-based Multi-agent Systems}

\author{%
  Hezhe Qiao\textsuperscript{1}, Hanghang Tong\textsuperscript{2}, Ee-Peng Lim\textsuperscript{1}, Bing Liu\textsuperscript{3}, Guansong Pang\textsuperscript{1}\thanks{Corresponding author: G. Pang} \\
  \textsuperscript{1}Singapore Management University\\
  \textsuperscript{2}University of Illinois at Urbana-Champaign  \\
  \textsuperscript{3}University of Illinois at Chicago \\
  \texttt{hezheqiao.2022@phdcs.smu.edu.sg} \\
  \texttt{htong@illinois.edu, liub@uic.edu} \\
  \texttt{\{eplim,gspang\}@smu.edu.sg} \\
   \\
}

\begin{document}

\maketitle
\vspace{-2.5em}
\begin{center}
\small
\href{https://hezheqiao2022.github.io/VerifyMAS/}{\faGlobe\ Project Page}
\quad\quad
\href{https://github.com/mala-lab/VerifyMAS}{\faGithub\ GitHub}
\end{center}

\begin{abstract}
Large language model-driven multi-agent systems (LLM-MAS) excel at complex tasks, yet unreliable agents remain a key bottleneck to system-level reliability. Automatic failure attribution is therefore critical, but existing approaches—such as direct prediction of agent–error pairs and agent-first failure attribution—rely on local logs of agent and miss global failures that only manifest over full interaction trajectories, such as cross-step inconsistencies and inter-agent coordination errors.
Moreover, directly predicting failures induces a large combinatorial search space, hindering fine-grained attribution. To address these challenges, we propose \textbf{VerifyMAS}, a hypothesis verification framework for agent failure attribution. Instead of directly predicting faulty agents and error types, VerifyMAS formulates and verifies failure hypotheses against full trajectories. This verification-based approach decomposes attribution into trajectory-level error validation and fine-grained agent localization, providing \textbf{an error-first attribution approach} that captures global failure patterns while substantially reducing the search space. We further introduce a hypothesis-based data construction strategy grounded in a structured error taxonomy and fine-tune a specialized LLM verifier model for trajectory-level failure verification and agent attribution. Experiments on \textit{Aegis-Bench} and \textit{Who\&When} show that VerifyMAS consistently improves diverse backbone models, including open-source Qwen and API-based GPT and Gemini models, outperforming prior methods without sacrificing inference efficiency for long multi-agent trajectories. Code and data are available at \renewcommand\UrlFont{\color{blue}}\url{https://github.com/mala-lab/VerifyMAS}.

\end{abstract}

\section{Introduction}
Large language model-driven multi-agent systems (LLM-MAS) have recently attracted significant attention due to their strong potential for enabling multiple agents to collaboratively solve complex tasks, such as reasoning, planning, decision-making, and tool use \cite{ung2022saferdialogues, du2024improving, pathak2025detecting, liu2026agentdog}. By decomposing a task into several subtasks and assigning them to specialized agents, LLM-MAS can often achieve better flexibility and efficiency than single-agent systems \cite{zhang2025g, zou2025latent}. However, this collaborative setting also introduces new challenges. Because the final outcome depends on interactions among multiple agents, failures in a single agent can propagate through the whole system, and eventually leading to incorrect or suboptimal outcomes.

In practice, agent failures can arise in various forms \cite{zhang2025graphtracer, cemri2025multi, zhu2025llm}. For example, an agent may deviate from task requirements, misinterpret its assigned role, overlook critical information from other agents, produce inconsistent reasoning, or terminate the process prematurely. These failures are often difficult to identify because they are often embedded within long interaction trajectories and may not be immediately observable in intermediate outputs \cite{koreeda2021contractnli, banerjee2025did}. Consequently, improving the reliability of LLM-MAS requires not only stronger collaboration strategies but also effective mechanisms for failure attribution \cite{cemri2025multi, zhu2025llm}.

\begin{wrapfigure}{r}{0.54\textwidth}
    \centering
    \includegraphics[width=\linewidth]{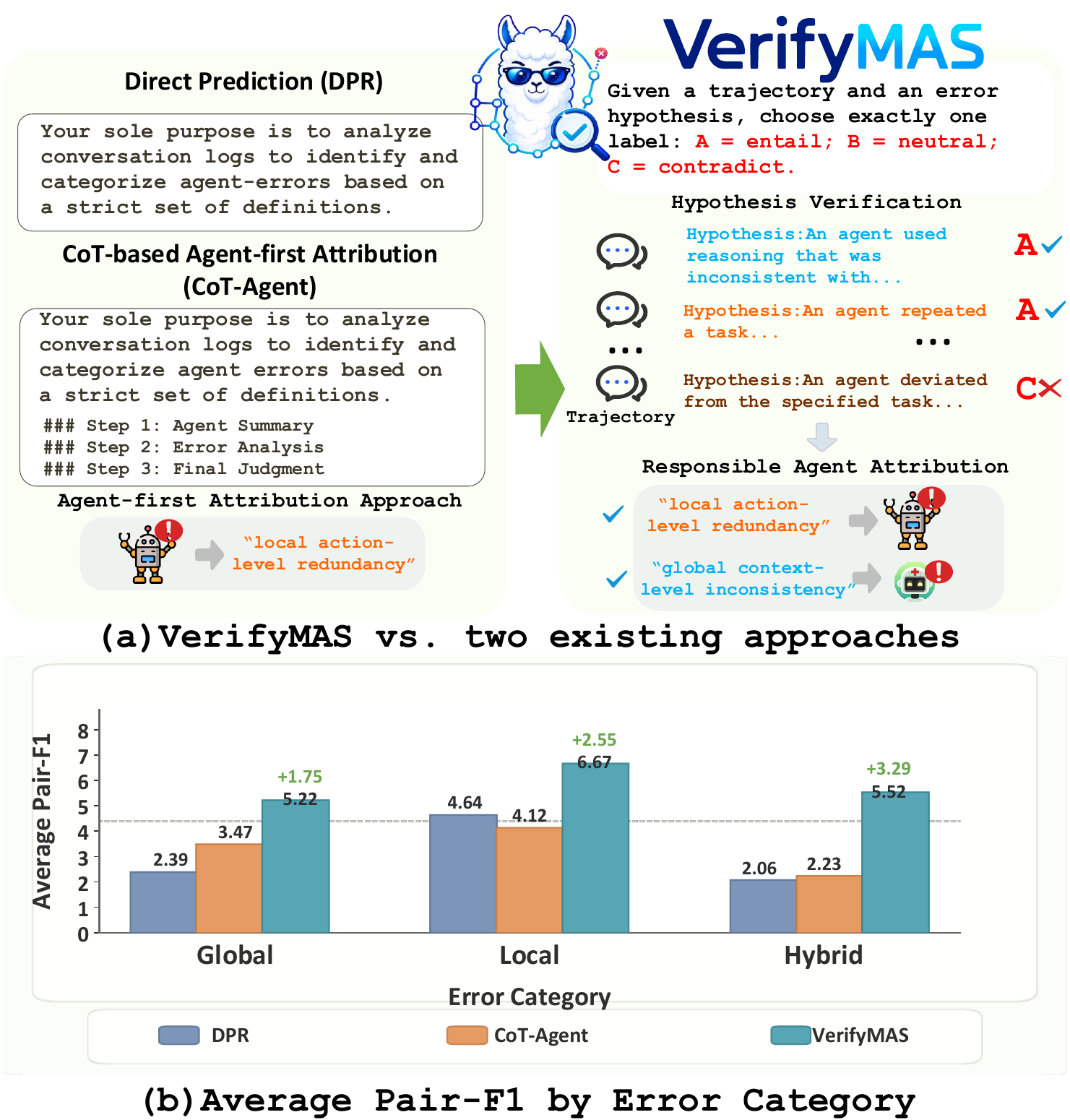}
    \caption{(a) Our hypothesis verification-based approach VerifyMAS vs. two existing approaches. (b) Performance of three approaches on Aegis-Bench \cite{kong2025aegis} in attributing three categories of error types—global, local, and hybrid errors in average Pair-F1.
    }
    \label{fig:example}
    \vspace{-1em}
\end{wrapfigure}

Failure attribution has emerged as a central task in guaranteeing the usefulness of LLM-MAS, focusing on identifying which agent caused a failure and classifying the types of errors that led to it.
Existing methods typically formulate it as a direct error-agent pair prediction problem, in which the predictions of error types and the faulty agent are combined into a single prediction task \cite{kong2025aegis, DBLP:conf/icml/ZhangY0LHZL0W0W25}, as shown in Fig. \ref{fig:example}\textbf{a}. While simplifying the problem, this formulation often i) overlooks explicit evidence grounding, ii) struggles to capture long-range dependencies across interaction steps, and iii) tends to produce surface-level error labels without sufficiently verifying whether the failure truly occurred or materially affected the final outcome. Moreover, this approach induces a large combinatorial search space of agents and failure categories, hindering fine-grained attribution.
Some recent work \cite{kong2025aegis, liu2026agentdog, DBLP:conf/icml/ZhangY0LHZL0W0W25} tackles this challenge using an agent-first attribution approach, which leverages chain-of-thought (CoT) prompting to first identify the responsible agent and then attribute the corresponding error, but this approach can easily bias the model toward local interaction context since each agent’s logs provide only partial observations of the full trajectory. 

To address these challenges, we propose \textbf{VerifyMAS}, a
hypothesis verification framework for agent failure attribution. The key idea is to augment LLMs with a set of hypotheses over predefined failure modes to automatically verify what type of error has occurred in a multi-agent trajectory and subsequently determine which agent is responsible. 
To this end, given a full trajectory, VerifyMAS verifies whether the trajectory entails, is neutral to, or contradicts each of the hypotheses describing the presence of an error type, yielding one of three possible outcomes: ``\textit{entail}'', ``\textit{neutral}'', and ``\textit{contradict}'', as shown in Fig.~\ref{fig:example}\textbf{a}. For entailed hypotheses, VerifyMAS further performs agent attribution to identify the responsible agent(s).
This decomposes the failure attribution into trajectory-level error validation and fine-grained faulty agent localization, thereby avoiding the combinatorial complexity from reasoning over the faulty agents and error types in a joint manner. 

VerifyMAS yields an error-first attribution methodology that prioritizes the verification of error types against the full trajectory, enabling accurate attribution of \textbf{global errors} that only manifest over the interaction log sequences of two or more agents, as shown in Fig.~\ref{fig:example}\textbf{b}. This contrasts fundamentally with existing agent-first methods that assess failures at the individual-agent level and are therefore prone to biases induced by local agent behaviors. Consequently, they often work well to attribute \textbf{local errors} that are observable at individual agents, but become ineffective to attribute global errors, having even worse performance in \textbf{hybrid errors} that require both agent-level local behavioral evidence and trajectory-level global context, as shown in Fig.~\ref{fig:example}\textbf{b} (see App. \ref{app:failure_type} for detailed examples of these error categories).
More importantly, 
our error-first approach VerifyMAS can also perform very well on attributing local and hybrid errors.
 In summary, this work makes the following three main contributions.


 
\begin{itemize}
\item We propose \textbf{VerifyMAS}, an error-first hypothesis verification framework for failure attribution in LLM-MAS. It decomposes failure attribution into error validation and fine-grained faulty agent localization, providing a principled solution for agentic failure attribution of both global and local errors. 

\item We propose a fine-tuning strategy tailored to the hypothesis verification approach, in which trajectory-level verification samples and agent-localization supervision are collected and leveraged to fine-tune an LLM verifier model under the VerifyMAS framework. This substantially enhances our model in failure diagnosis of in-distribution trajectories while preserving robust generalization to out-of-distribution trajectories. The dataset will be released to promote more advances in this line.

\item Extensive experiments on \textit{Aegis-Bench} and \textit{Who\&When} demonstrate that VerifyMAS consistently improves diverse open-source and proprietary models. We further validate its effectiveness under the SFT setting, where hypothesis-verification-based fine-tuning strengthens in-distribution diagnostic ability while preserving out-of-distribution generalization.

\end{itemize}
    
\section{Related Work}

\paragraph{Safeguarding MAS}
LLM-based guard models have recently been studied as lightweight classifiers for monitoring and filtering model misbehaviors ~\cite{wang2025g, han2024wildguard,kwon2024slm, ghosh2025aegis2}.  Recent work extends this idea to agentic systems, where evaluators need to inspect intermediate steps rather than only final answers~\cite{zhuge2025agent, xiang2025guardagent, jia2026mas, zheng2026rethinking}.
Early representative systems, such as Llama Guard \cite{inan2023llama}, ShieldGemma \cite{zeng2024shieldgemma}, and Aegis \cite{ghosh2025aegis2} mainly focus on content safety moderation, where a guard model classifies user inputs or model responses according to a predefined risk taxonomy.
These methods demonstrate that instruction-tuned LLMs can serve as flexible taxonomy-guided classifiers.  Different from these guard models that perform content-level safety classification of isolated prompts or responses, our setting focuses on the analysis of complete multi-agent interaction trajectories to attribute failures. 

Another line of work studies guard models from the perspective of anomaly detection \cite{qiao2025deep, zhou2025guardian, pan2025explainable, advani2026trajectory}, aiming to detect behaviors (\eg, failures) that deviate from normal or expected MAS patterns. The task is orthogonal to failure attribution that provides detailed root cause analysis of the detected failures. Additionally, they focus on learning representation-level deviation patterns, while our setting 
require contextual semantic understanding of the tasks assigned to MAS and their full trajectory.



\paragraph{MAS Failure Attribution}
Failure attribution has recently emerged as a critical task for safeguarding the reliability of LLM-MAS. Early studies~\cite{cemri2025multi, kong2025aegis, DBLP:conf/icml/ZhangY0LHZL0W0W25, zhu2025llm, liu2026agentdog} introduce predefined taxonomies of failure modes and formulate failure attribution as direct agent-error pair prediction, jointly determining both the error type and the responsible agent within a single prediction task.  However, directly predicting agent-error pairs introduces a large combinatorial search space are struggle to capture long-range dependencies. Some recent works \cite{kong2025aegis, DBLP:conf/icml/ZhangY0LHZL0W0W25,liu2026agentdog, DBLP:journals/corr/abs-2509-03312} further leverage chain-of-thought prompting to improve failure attribution by first analyzing the behaviors of each agent and then identifying possible failures, forming an agent-first attribution approach. Although such methods can provide more explicit intermediate analysis compared with direct prediction, starting from individual agents may bias the model toward local action-level mistakes, while overlooking trajectory-level evidence that emerges only from cross-step interactions or coordination failures \cite{turpin2023language, jia2026mas, cemri2025multi, kong2025aegis}.

 
Different from prior work that directly predicts the responsible agent-error pair and agent-first attribution approach, we formulate agent failure attribution as a grounded hypothesis-verification problem over long multi-agent trajectories, enabling finer-grained reasoning over specific error-agent types and explicit support/contradiction signals from the whole trajectory evidence with global context instead of local agent-level behavioral evidence.

\section{Our Approach: VerifyMAS}

\paragraph{Problem Statement}
An LLM-MAS interaction trajectory is defined as $\tau=(I,l_1,\ldots,l_T)$, where $I$ represents the initial task instruction, 
$l_1, l_2, \cdots, l_T$ denote a collection of log sequences produced by a finite set of $K$ LLM agents $\mathcal{A}=\{a_1,a_2,\ldots,a_K\}$ over $T$ interaction steps to solve a given task, with each $l_i$ denoting a log sequence that traces the actions and reasoning process of a specialized agent $a_j$ taken at step $i$. An agent can be assigned to work in different steps, so we typically have $T \gg K$.
Given a trajectory of an LLM-MAS failure $\tau$, our objective is to attribute the failure by determining which agent caused a specific error that led to the failure of the overall agent system.
The error space is defined by a predefined taxonomy of $M$ distinct error types, denoted as $\mathcal{Y}=\{y_1,\ldots,y_M\}$. For a failure trajectory, the ground-truth annotation is represented as a structured set of agent--error pairs, denoted by $\mathcal{G}(\tau)=\{(a_1^{*},y_1^{*}),(a_2^{*},y_2^{*}),\ldots \}$, where $y^{*}_i \in \mathcal{Y}$ denotes the error mode and $a^{*}_j$ denotes the agent responsible for causing the error. A single trajectory can contain either single or multiple agent--error pairs as its ground truth, \ie, one trajectory may have multiple labels.
The core task is to obtain a diagnostic function $f_\theta$ that maps a trajectory $\tau$ to a predicted attribution set that approximates the ground truth:
$f_\theta:\tau\mapsto\hat{\mathcal{G}}(\tau)$.





\begin{figure*}
    \centering
    \includegraphics[width=\linewidth]{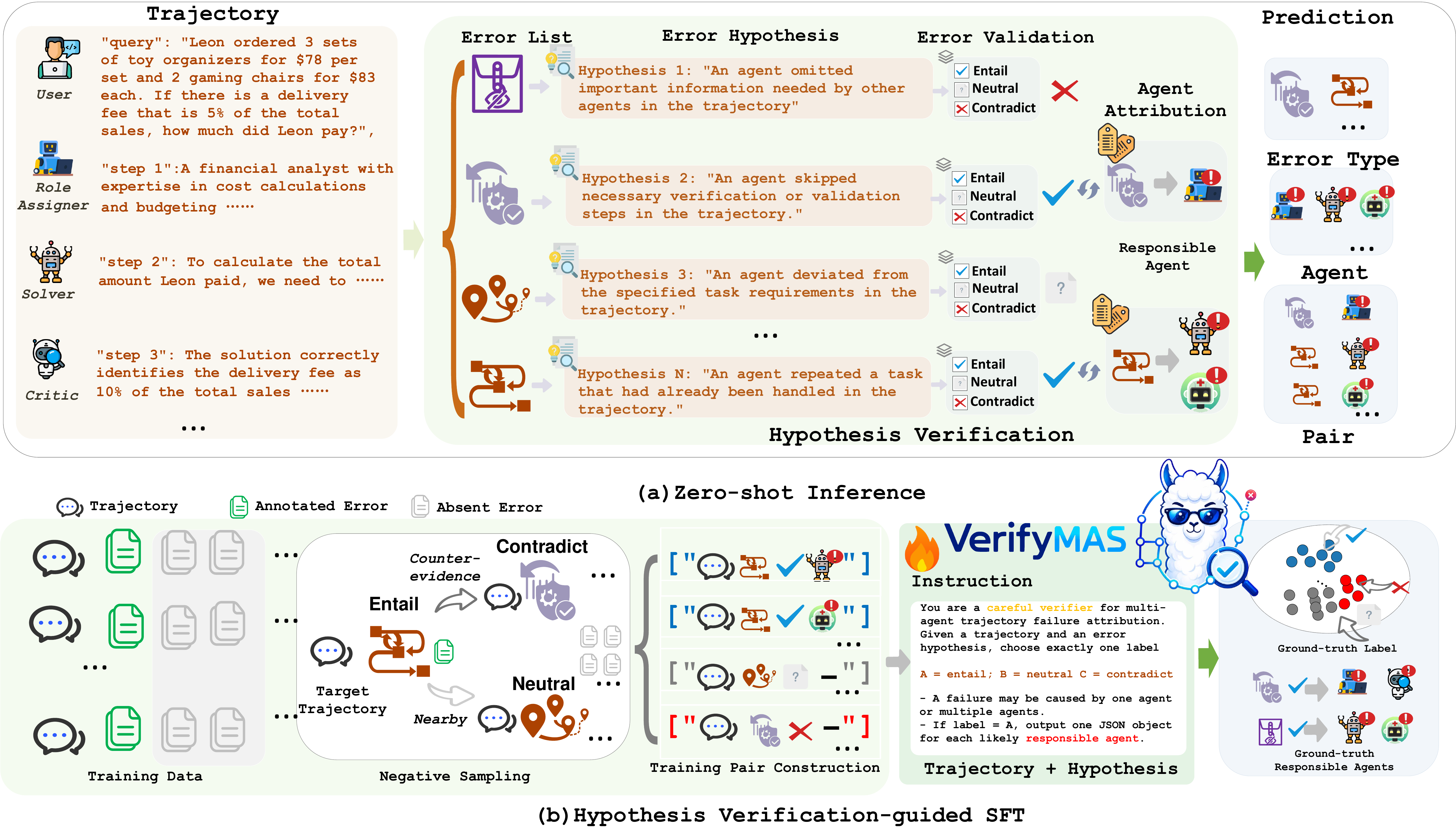}
    \caption{Overview of the proposed VerifyMAS. \textbf{(a)} Zero-shot inference. A trajectory is paired with hypotheses on predefined error types, and then an LLM predicts whether the trajectory is “entail”, “neutral”, or “contradict” w.r.t. each hypothesis describing the presence of an error type. The entailed hypotheses are further examined for faulty agent attribution, producing the final error–agent predictions. \textbf{(b)} Hypothesis verification-guided supervised fine-tuning. We construct trajectory–hypothesis pairs by pairing each trajectory with both annotated errors and absent errors. VerifyMAS is then fine-tuned to jointly predict the type of the error and the responsible agent.}
    \label{fig:framework} 
   
\end{figure*}

\paragraph{Overview of VerifyMAS}
As shown in Fig. \ref{fig:framework}, our proposed VerifyMAS framework can operate under either zero-shot inference or supervised fine-tuning (SFT). In the zero-shot inference setting, given a multi-agent trajectory, we first instantiate a set of hypotheses based on the predefined failure taxonomy. Each hypothesis is paired with the trajectory and evaluated by an LLM verifier, which predicts one of three labels: entail, neutral, or contradict. Hypotheses that are labeled as entail are retained as candidate failures, upon which the model further performs faulty agent attribution, identifying the most likely responsible agent for each supported error hypothesis. In SFT,  we leverage the trajectories with annotated faulty agents and error types to construct training instances by pairing each trajectory with error hypotheses and assigning labels. For an entailed hypothesis, the corresponding responsible agents are used as supervision for agent attribution, while neutral and contradictory hypotheses are assigned an empty agent label. These trajectory–hypothesis pairs, together with their error type and agent annotations, are then used to perform SFT of our LLM verifier, so that the verifier learns to jointly verify hypotheses and attribute responsibility to the faulty agent.

\subsection{Zero-shot Failure Attribution}
\paragraph{Error Hypothesis Formulation}
Instead of directly predicting multiple error labels from the entire trajectory, we reformulate agent failure attribution as a grounded hypothesis verification problem. This enforce the model to skip the local context of the agent and evaluate the trajectory from a global perspective. We construct several natural-language hypotheses that only ask whether a specific error type $y_m \in \mathcal{Y}$ is supported by the trajectory, without assigning it to any particular agent. Formally, for each error type $y_m$, we construct a hypothesis $h_m$ describing that some agent in the trajectory exhibits error type $y_m$. For example, for the error type \emph{task specification deviation}, the corresponding hypothesis is instantiated as: \textit{An agent deviated from the specified task requirements in the trajectory}.

\paragraph{Error-hypothesis Verification}
Based on the above formulation, we perform failure attribution in two stages. In the first stage, we verify all type-level hypotheses $\{h_m\}_{y_m \in \mathcal{Y}}$ against the trajectory to identify candidate error types. This stage is designed to identify failure modes that can only be reliably recognized from global trajectory-level evidence rather than directly traversing agents and judging errors from local agent logs. This is more suitable for detecting errors that emerge from cross-step dependencies, long-range context, or multi-agent interactions. Formally, we formulate this step as a three-way hypothesis validation problem, where the predicted label is selected from
\[
z_{m} \in \{A, B, C\},
\]
with $A$, $B$, and $C$ denoting \textit{entail}, \textit{neutral}, and \textit{contradict}, respectively. Specifically, $A$ indicates that the trajectory provides clear evidence for the hypothesized failure, $B$ indicates that the evidence is insufficient or ambiguous, and $C$ indicates that the hypothesis is contradicted by the trajectory. We adopt a three-way formulation rather than a binary one because many failure hypotheses cannot be confidently verified or rejected from the trajectory alone. The \textit{neutral} class captures cases with weak, incomplete, or ambiguous evidence, as well as errors whose impact on the final outcome is unclear. This avoids forcing uncertain cases into either positive or negative labels and leads to more reliable failure verification.
Formally, given a trajectory $\tau_i$ and hypothesis $h_m$ for error type $y_m$, the verifier predicts the label for error $\hat{z}_{i,m}$ by
\[
\hat{z}_{i,m}
=
\arg\max_{z \in \{A,B,C\}}
f_\theta(z \mid \tau_i, h_m, P, \mathcal{A}_i),
\]
where $P$ denotes the instruction prompt that guides the LLM to verify whether the given hypothesis is supported by the trajectory $\tau_i$, and $\mathcal{A}_i$ is the candidate agent set for $\tau_i$. Only hypotheses predicted as entail are passed to the agent attribution stage. Therefore, the retained failures for $\tau_i$ are 
\[
\widehat{\mathcal{Y}}_i
=
\{y_{m} \mid \hat{z}_{i,m}=A\},
\]
where $\widehat{\mathcal{Y}}_i$ are selected error modes from all the error-hypothesis verification for trajectory $\tau$.


\paragraph{Agent-level Attribution}
After hypothesis validation, agent attribution is performed only for hypotheses with label $A$. 
Note that different trajectories may contain different numbers and names of agents.
For each retained error type $y_m \in \widehat{\mathcal{Y}}$, the model identifies which candidate agents are responsible for the entail failure. The responsible agents must be selected from the current candidate set $\mathcal{A}_i$, rather than from a fixed global agent vocabulary.

For cases where multiple agents are responsible for the same entail failure, the model outputs one JSON object for each responsible agent, with each object containing exactly one agent. The final attributed failure set for trajectory $\tau_i$ is then written as
\[
\widehat{\mathcal{G}}(\tau_i)
=
\{(a_n,y_m) \mid y_m \in \mathcal{Y},\ a_n \in \mathcal{A}_i\},
\]
where $\widehat{\mathcal{G}}(\tau_i)$ denotes the predicted responsible agent-error pair where each tuple $(a_n,y_m)$ indicates that agent $a_n$ is inferred to be responsible for the entail error type $y_m$.

This two-stage design separates failure validation from agent localization. Our key insight is that many failure modes are not evident from a single agent’s local log, but only emerge when considering the full trajectory, including cross-step dependencies, inter-agent interactions, and the final task outcome. Therefore, we first start from the error side and verify whether each hypothesized failure is supported by the global trajectory context.  Meanwhile, conditioning agent attribution on validated failures allows the model to focus on agents whose actions are relevant to the detected error, making it easier to identify problematic inter-agent interactions.


\subsection{Hypothesis Verification-guided Supervised Fine-tuning}

\noindent \textbf{Training Data Construction} We construct the supervised fine-tuning dataset from annotated multi-agent trajectories. 
Each trajectory is associated with ground-truth faulty agents and their corresponding error types. For each trajectory, we first generate entailment examples using the ground-truth annotations. Specifically, if an agent $a$ is annotated as responsible for an error type $y_m$, given the input ${{\bf{x}}_{i,m}}= \{\tau_i, h_m, P, \mathcal{A}_i\}$ with  trajectory corresponding to error hypothesis $h_m$ , instruction prompt $P$, and the candidate agent list $\mathcal{A}_i$ in the trajectory, we construct a positive example with label $entail$ as the corresponding output:
\[
{\bf y}_{i,m} = \{\texttt{"label"}:\texttt{"A"},  \texttt{"agent"}:[a]\}, \hat{z}_{i,m}=A,
\]

where $a \in \mathcal{A}_i$ cause the error $y$ in the trajectory. 
If multiple agents are responsible for the same entailed error hypothesis, we construct multiple JSON objects, each corresponding to one responsible agent and labeled as entail. These positive examples teach the model to recognize when a hypothesized failure is supported by the trajectory and to attribute it to the responsible agent. 

To construct contradict and neutral examples, we sample errors from absent error \ie ${\mathcal{Y}_a=\mathcal{Y}/\mathcal{Y}_e}$ where $\mathcal{Y}_e$ is the set of existing annotated errors, to formulate the negative samples for this trajectory. 
For contradicted examples, we sample errors from absent error types that are explicitly refuted by trajectory evidence and semantically incompatible with the ground-truth failure annotation to formulate the hypotheses.
Specifically, we construct an explicit counter-evidence table for each error type, see App.~\ref{app:failure_mapping}, which specifies evidence cues whose presence in the trajectory indicates that the corresponding error is unlikely to have occurred. Therefore, when such cues are observed, the associated absent-error hypothesis can be treated as a contradicted hypothesis. 

For neutral examples, we sample absent error types following a nearby error-type mapping table, see App.~\ref{app:failure_mapping}, which groups semantically related or easily confusable error types according to their definitions and behavioral features. As a result, neutral examples often correspond to partially plausible agent behaviors or related failure modes for which the trajectory provides insufficient evidence to make a confident entailment decision. These two types of negative samples provide fine-grained negative supervision that helps the verifier distinguish clearly contradicted hypotheses from plausible but unsupported neutral hypotheses.
The corresponding outputs for the contradict ${\bf{x}}_{contradict}$ and neutral ${\bf{x}}_{neutral}$ input with hypothesis $h_m$ are 
\[
\begin{cases}
{\bf{y}}_{\text{neutral}} = \{\texttt{"label"}:\texttt{"B"},\texttt{"agents"}:[]\}, 
& \hat{z}=B,\\
{\bf{y}}_{\text{contradict}}= \{\texttt{"label"}:\texttt{"C"},\texttt{"agents"}:[]\}, 
& \hat{z}=C.
\end{cases}
\]

\noindent \textbf{Training Loss Function}
We formulate the hypothesis-verification approach based SFT as a unified structured-output generation problem. For each instance ${\bf{x}}_{i,m}$, which contains the trajectory $\tau_i$, the error hypothesis $h_m$,  the candidate agent set $\mathcal{A}_i$, and the instruction $P$, the model is trained to generate a serialized JSONL-style response.
The target response $\mathbf{y}_{i,m}$ contains both the verification label and the responsible agents.

When multiple agents are responsible, the target is serialized as multiple JSON objects, one for each responsible agent; when the hypothesis is neutral or contradict, the target is a single JSON object with an empty agent list.
This design naturally supports instance-specific candidate agent sets and avoids introducing a fixed global agent classifier. Specifically, given the input ${\bf{x}}_{i,m}$
, after serializing the target JSON object(s) into a response string and tokenizing it, we denote the gold response as
$\mathbf{y}_{i,m}^*=(y_{i,m}^{(1)*}, y_{i,m}^{(2)*}, \ldots, y_{i,m}^{(|\mathbf{y}_{i,m}^*|)*})$,
where $y_{i,t}^*$ denotes the $t$-th token in the serialized target response and
$|\mathbf{y}_{i,m}^*|$ is the number of response tokens.
The training loss $\mathcal{L}_{\mathrm{HSFT}}$ for ${\bf x}_{i,m}$ is

\[
\mathcal{L}_{\mathrm{HSFT}}
=
-
\sum_{t=1}^{|\mathbf{y}_{i,m}^*|}
\log p_{\theta}
\left(
y_{i,m}^{(t)*}
\mid
{\bf x}_{i,m}, y_{i,m}^{(<t)*}
\right),
\]
In this way, the model is supervised to generate structured decisions that first verify whether the hypothesis is entailed and then output the corresponding responsible agent(s) when applicable.



\section{Main Experiments}
\noindent \textbf{Datasets.} 
Following previous work \cite{kong2025aegis, DBLP:conf/icml/ZhangY0LHZL0W0W25}, we evaluate all methods on two benchmark datasets: Aegis-Bench \cite{kong2025aegis} and Who\&When \cite{DBLP:conf/icml/ZhangY0LHZL0W0W25}. 
Aegis-Bench \cite{kong2025aegis} is a trajectory dataset that was collected from six task domains and six multi-agent system frameworks. Who\&When \cite{DBLP:conf/icml/ZhangY0LHZL0W0W25} is a failure log dataset collected from 127 LLM multi-agent systems with fine-grained annotations linking failures to specific agents and decisive error steps. It is used as an out-of-distribution (OOD) benchmark for evaluation.  Details information about datasets can be found in App. \ref{app:datasets}.


\noindent  \textbf{Evaluation Metrics.} 
Following prior work \cite{kong2025aegis, DBLP:conf/icml/ZhangY0LHZL0W0W25},  we employ Micro-F1 ($\mu$F1) and Class-wise Macro-F1 (MF1) as the primary metrics. In particular,$\mu$F1 emphasizes overall classification performance, whereas MF1 highlights performance across different classes by giving equal weight to each class. Because a single trajectory may involve multiple faulty agents and multiple error modes, we evaluate performance at three levels of granularity: \textbf{Pair}-level, which measures correctly identified agent–error pairs; \textbf{Agent}-level, which measures faulty-agent identification regardless of error type; and \textbf{Error}-level, which measures error-mode identification regardless of the responsible agent. In addition to F1, recall and precision are also provided as auxiliary metrics.

\noindent  \textbf{Competing Methods and Implementation Details.} 
We compare VerifyMAS with not-formally-published approaches addressing closely related problems, AgentDebug \citep{zhu2025llm} and AgentDoG \citep{liu2026agentdog} and a range of open-source models, including Qwen models at different scales: 7B-Instruct, 8B, 9B in non-thinking, and 14B-Instruct \cite{DBLP:journals/corr/abs-2412-15115, qwen3technicalreport, team2026qwen3}. We also compare with Proprietary models, including GPT-4.1 \citep{achiam2023gpt}, GPT-4o-mini \citep{hurst2024gpt},  Gemini-2.5-Flash \citep{comanici2025gemini}, and Gemini-2.5-Pro  \citep{comanici2025gemini}. All these models are evaluated under a zero-shot setting with the instruction prompt provided in Aegis \cite{kong2025aegis}. We implement VerifyMAS over the base model, with the detailed instruction prompt provided in the App. \ref{app:prompts}. The same task instructions are used in our SFT models.  In addition, we compare our SFT method against the contrastive-based DCL model \cite{kong2025aegis} and several open-source baselines under SFT setting \cite{DBLP:journals/corr/abs-2412-15115, qwen3technicalreport}.

\noindent  \textbf{Training Details.} 
 For our SFT experiments, we utilize the verl library \cite{sheng2025hybridflow}, and all fine-tuning is conducted on a cluster of 2 NVIDIA H200 GPUs. The learning rate is set to $1e-4$, and the batch size is set to 64. VerifyMAS is trained for 2 epochs based on Qwen2.5-7B-Instruct and Qwen3-8B. More details of hyperparameters for all SFT runs can be found in the App. \ref{app:training_details}.

\definecolor{GroupGray}{RGB}{242,242,242}
\definecolor{UpColor}{RGB}{250,95,35}
\definecolor{DownColor}{RGB}{90,105,125}

\newlength{\mainmetricw}
\setlength{\mainmetricw}{4.25em}

\newcolumntype{M}{>{\centering\arraybackslash}p{\mainmetricw}}

\providecommand{\ourup}[1]{}
\providecommand{\ourdown}[1]{}

\renewcommand{\ourup}[1]{%
  \makebox[0pt][l]{%
    \smash{\textsuperscript{%
      \tiny\bfseries\textcolor{UpColor}{$\uparrow$#1}%
    }}%
  }%
}

\renewcommand{\ourdown}[1]{%
  \makebox[0pt][l]{%
    \smash{\textsuperscript{%
      \tiny\bfseries\textcolor{DownColor}{$\downarrow$#1}%
    }}%
  }%
}
\begin{table}[t]
\centering
\captionsetup{skip=6pt}
\fontsize{8.8pt}{9.2pt}\selectfont
\setlength{\tabcolsep}{0.25pt}
\renewcommand{\arraystretch}{0.92}

\caption{Performance gains of VerifyMAS-enabled models over their corresponding base models. The best performance is boldfaced. Arrows indicate changes over the base model, where \textcolor{UpColor}{$\uparrow$} denotes improvement and \textcolor{DownColor}{$\downarrow$} denotes degradation.}
\label{tab:main}

\resizebox{\linewidth}{!}{%
\begin{tabular}{@{}l*{12}{M}@{}}
\toprule
\multirow{3}{*}{\textbf{Model}} 
& \multicolumn{6}{c}{\textbf{Aegis-Bench}} 
& \multicolumn{6}{c}{\textbf{Who\&When}} \\
\cmidrule(lr){2-7} \cmidrule(lr){8-13}

& \multicolumn{2}{c}{\textbf{Pair}} 
& \multicolumn{2}{c}{\textbf{Agent}} 
& \multicolumn{2}{c}{\textbf{Error}}
& \multicolumn{2}{c}{\textbf{Pair}} 
& \multicolumn{2}{c}{\textbf{Agent}} 
& \multicolumn{2}{c}{\textbf{Error}} \\
\cmidrule(lr){2-3} \cmidrule(lr){4-5} \cmidrule(lr){6-7}
\cmidrule(lr){8-9} \cmidrule(lr){10-11} \cmidrule(lr){12-13}

& $\mu$F1 & MF1 
& $\mu$F1 & MF1 
& $\mu$F1 & MF1 
& $\mu$F1 & MF1 
& $\mu$F1 & MF1 
& $\mu$F1 & MF1 \\

\midrule
\rowcolor{GroupGray}
\multicolumn{13}{c}{\textbf{\small Open-source Models}} \\
AgentDebug* 
& 4.34 & 0.49
& 45.90 & \textbf{16.29}
& 11.53 & 2.25
& 3.47 & 1.81
& 37.45 & 33.62
& 9.25 & 1.54 \\

AgentDoG* 
& 0.83 & 0.11
& 13.01 & 4.07
& 3.34 & 1.55
& 0.90 & 0.25
& 12.67 & 9.13
& 3.70 & 2.45 \\
\midrule
Qwen2.5-7B-Instruct      
& 5.02 & 2.52 & 27.55 & 14.49 & 14.96 & 11.36 
& 2.31 & 1.14 & 40.92 & 23.50 & 3.64 & 1.77 \\

\quad +VerifyMAS   
& 5.89\ourup{0.87} 
& \textbf{2.98}\ourup{0.46} 
& 47.92\ourup{20.37} 
& 14.66\ourup{0.17} 
& 25.07\ourup{10.11} 
& 24.26\ourup{12.90} 
& 5.83\ourup{3.52} 
& \textbf{3.44}\ourup{2.30} 
& 39.26\ourdown{1.66} 
& 33.50\ourup{10.00} 
& 13.33\ourup{9.69} 
& 9.54\ourup{7.77} \\

\midrule

Qwen3-8B-Non-Thinking               
& 3.96 & 1.40 & 21.34 & 8.16 & 15.81 & 13.89 
& 3.88 & 1.81 & 27.78 & 17.64 & 3.88 & 1.91 \\

\quad +VerifyMAS   
& 5.49\ourup{1.53} 
& 1.85\ourup{0.45} 
& 42.96\ourup{21.62} 
& 11.65\ourup{3.49} 
& 26.50\ourup{10.69} 
& 24.05\ourup{10.16} 
& 4.92\ourup{1.04} 
& 2.88\ourup{1.07} 
& 35.03\ourup{7.25} 
& 24.11\ourup{6.47} 
& 11.93\ourup{8.05} 
& 5.48\ourup{3.57} \\
\hline
Qwen2.5-14B-Instruct     
& 5.47 & 2.20 & 35.78 & 12.71 & 20.24 & 5.91  
& 0.00 & 0.00 & \textbf{49.88} & 33.19 & 1.56 & 1.35 \\

\quad +VerifyMAS  
& \textbf{7.40}\ourup{1.93} 
& 2.69\ourup{0.49} 
& \textbf{48.95}\ourup{13.17} 
& 14.50\ourup{1.79} 
& \textbf{28.37}\ourup{8.13} 
& \textbf{26.90}\ourup{20.99} 
& 5.15\ourup{5.15} 
& 2.68\ourup{2.68} 
& 35.97\ourdown{13.91} 
& 27.53\ourdown{5.66} 
& 10.53\ourup{8.97} 
& 7.04\ourup{5.69} \\
\hline
Qwen3.5-9B-Non-Thinking  & 3.35 & 1.02 & 25.21 & 9.26 & 10.73 & 8.63
& 5.18 & 2.86 & 45.28 & 35.60 & 9.73 & 4.33 \\
\quad +VerifyMAS
& 5.51\ourup{2.16}
& 2.18\ourup{1.16}
& 41.65\ourup{16.44}
& 15.81\ourup{6.55}
& 27.15\ourup{16.42}
& 26.30\ourup{17.67}
& \textbf{6.16}\ourup{0.98}
& 3.01\ourup{0.15}
& 46.63\ourup{1.35}
& \textbf{36.42}\ourup{0.82}
& \textbf{15.45}\ourup{5.72}
& \textbf{10.74}\ourup{6.41}
\\
\midrule

\rowcolor{GroupGray}
\multicolumn{13}{c}{\textbf{\small Proprietary Models}} \\
AgentDebug**
& 5.41 & 0.58
& 44.01 & 14.71
& 12.64 & 2.68
& 4.57 & 2.68
& 44.70 & 38.04
& 9.30 & 4.31 \\

AgentDoG**
& 6.19 & 0.93
& 52.01 & 17.94
& 14.69 & 5.80
& 5.00 & 2.62
& 43.05 & 37.52
& \textbf{15.15} & 6.55 \\
\midrule
GPT-4o-mini      
& 5.76 & 1.63 & 38.54 & 14.72 & 19.95 & 16.02 
& 2.11 & 0.98 & 47.42 & 34.21 & 5.26 & 3.33 \\

\quad +VerifyMAS 
& 6.58\ourup{0.82} 
& 2.62\ourup{0.99} 
& 49.30\ourup{10.76} 
& 16.54\ourup{1.82} 
& 27.89\ourup{7.94} 
& \textbf{27.79}\ourup{11.77} 
& 5.19\ourup{3.08} 
& 3.19\ourup{2.21} 
& 44.96\ourdown{2.46} 
& 38.00\ourup{3.79} 
& 12.39\ourup{7.13} 
& 11.36\ourup{8.03} \\

\midrule

GPT-4.1          
& 7.44 & 2.27 & 37.48 & 11.12 & 20.65 & 15.75 
& 3.36 & 1.16 & 42.29 & 28.93 & 7.00 & 5.84 \\

\quad +VerifyMAS 
& \textbf{7.59}\ourup{0.15} 
& 2.79\ourup{0.52} 
& 49.52\ourup{12.04} 
& 17.07\ourup{5.95} 
& \textbf{27.95}\ourup{7.30} 
& 27.37\ourup{11.62} 
& 5.09\ourup{1.73} 
& 2.83\ourup{1.67} 
& 42.93\ourup{0.64} 
& 37.44\ourup{8.51} 
& 13.06\ourup{6.06} 
& 10.66\ourup{4.82} \\

\midrule

Gemini-2.5-Flash 
& 6.99 & 2.76 & 42.02 & 16.45 & 23.47 & 19.85 
& 7.32 & 3.33 & \textbf{55.56} & 36.98 & 11.94 & 7.96 \\

\quad +VerifyMAS 
& 7.02\ourup{0.03} 
& 2.77\ourup{0.01} 
& 50.33\ourup{8.31} 
& 16.91\ourup{0.46} 
& 27.23\ourup{3.76} 
& 26.73\ourup{6.88} 
& \textbf{7.39}\ourup{0.07} 
& 3.49\ourup{0.16} 
& 43.25\ourdown{12.31} 
& 37.23\ourup{0.25} 
& 13.33\ourup{1.39} 
& \textbf{11.65}\ourup{3.69} \\
\midrule
Gemini-2.5-Pro   & 6.96 & \textbf{2.88} & 41.32 & 16.15 & 19.93 & 16.29 & 6.81 & 2.69 & 53.11 & 34.92 & 11.07 & 8.11  \\
\quad +VerifyMAS & 7.47\ourup{0.51} & 2.60\ourdown{0.28} & \textbf{53.61}\ourup{12.29} & \textbf{18.21}\ourup{2.06} & 27.27\ourup{7.34} & 26.99\ourup{10.70} & 7.33\ourup{0.52} & \textbf{4.18}\ourup{1.49} & 49.10\ourdown{4.01} & \textbf{41.47}\ourup{6.55} & 12.94\ourup{1.87} & 11.39\ourup{3.28}\\
\bottomrule
\end{tabular}%
}
\end{table}

\subsection{Zero-shot Results}

To examine the effect of VerifyMAS, we first implement VerifyMAS using several open-source models and proprietary models, then we conduct a zero-shot evaluation on the AEGIS and Who\&When benchmarks, comparing it against their base models. The main results are shown in Table \ref{tab:main}.  The detailed recall and precision results can be found in the App. \ref{app:additional}.
From the results, (1) we observe that applying VerifyMAS yields a substantial absolute gain compare the original open-source model. 
On Aegis-Bench, VerifyMAS can consistently improve performance across Pair, Agent, and Error levels, with particularly large gains on Agent $\mu$F1 and Error MF1. This indicates that starting from the error hypothesis verification enables the model to better capture global trajectory-level errors and provides a more reliable basis for subsequent agent localization. On the Who\&When benchmark, VerifyMAS leads to a slight drop in agent-level identification on 7B-Instruct and 14B-Instruct, while an improvement was observed on the 8B model, mainly because different models vary in their ability to attribute verified failures to the correct agents. Overall, VerifyMAS consistently improves the overall average score at the pair and failure levels across multiple Qwen models, indicating stronger generalization to unseen trajectory distributions. (2) VerifyMAS can also benefit stronger proprietary models. These results show that applying VerifyMAS to proprietary models like GPT-4.1 and Gemini-2.5-Pro can improve the average score across three levels.  This further demonstrates that  VerifyMAS is model-agnostic and can consistently enhance both open-source and closed-source models. The gains are especially pronounced at the Error level, suggesting that hypothesis-guided trajectory-level verification is effective for detecting failure modes that require global context, cross-step dependencies, and inter-agent interaction evidence. 


\noindent \textbf{Comparison with AgentDoG and AgentDebug.}
To further demonstrate the effectiveness of VerifyMAS, we compare VerifyMAS with two concurrent, not-formally-published approaches addressing closely related problems, AgentDebug \citep{zhu2025llm} and AgentDoG \citep{liu2026agentdog}. We adapted both methods to our evaluation setting and included them as additional SOTA baselines.
For a comprehensive and fair comparison, we implemented both baselines with Qwen2.5-7B (AgentDebug* and AgentDoG*) and GPT-4o-mini (AgentDebug** and AgentDoG**). We then compared each baseline against the corresponding VerifyMAS variant using the same backbone model. The results show that VerifyMAS (with either Qwen2.5-7B or GPT-4o-mini) significantly outperforms both AgentDebug and AgentDoG at both the error-type and agent–error pair levels on the two datasets, e.g., achieving up to 5.0+ (Pair $\mu$F1) and 20.0+ (Error $\mu$F1/MF1) improvement. This advantage stems from the fact that these two baselines are primarily designed for step-level or agent-level failure detection rather than fine-grained failure attribution. Although the error-first principle renders VerifyMAS less advantageous in agent-level performance, benefiting from its progressive and stringent verification, VerifyMAS remains highly competitive to these two methods in both agent-level metrics.

\subsection{Supervised Fine-tuning Results}
To further demonstrate the effectiveness of our approach, we conduct experiments under the SFT setting on the Aegis-Bench and Who\&When benchmarks. In this setting, VerifyMAS is trained on the constructed error hypothesis verification training data and compared with local Qwen-based models.

\definecolor{GroupGray}{RGB}{242,242,242}
\definecolor{UpColor}{RGB}{250,95,35}
\definecolor{DownColor}{RGB}{90,105,125}

\newlength{\sftmetricw}
\setlength{\sftmetricw}{3.85em}

\newcolumntype{T}{>{\centering\arraybackslash}p{\sftmetricw}}

\newcommand{\sftup}[1]{%
  \makebox[0pt][l]{%
    \smash{\textsuperscript{%
      \scriptsize\bfseries\textcolor{UpColor}{$\uparrow$#1}%
    }}%
  }%
}

\newcommand{\sftdown}[1]{%
  \makebox[0pt][l]{%
    \smash{\textsuperscript{%
      \scriptsize\bfseries\textcolor{DownColor}{$\downarrow$#1}%
    }}%
  }%
}

\begin{wraptable}{r}{0.59\textwidth}
\centering
\fontsize{7.8pt}{8.2pt}\selectfont
\setlength{\tabcolsep}{0.3pt}
\renewcommand{\arraystretch}{0.90}

\caption{SFT performance comparison.}
\label{tab:sft_results}

\resizebox{\linewidth}{!}{%
\begin{tabular}{@{}l*{6}{T}@{}}
\toprule
\multirow{2}{*}{\textbf{Model}}
& \multicolumn{2}{c}{\textbf{Pair}}
& \multicolumn{2}{c}{\textbf{Agent}}
& \multicolumn{2}{c}{\textbf{Error}} \\
\cmidrule(lr){2-3} 
\cmidrule(lr){4-5} 
\cmidrule(lr){6-7}
& $\mu$F1 & MF1
& $\mu$F1 & MF1
& $\mu$F1 & MF1 \\
\midrule

\rowcolor{GroupGray}
\multicolumn{7}{c}{\textbf{Aegis-Bench}} \\

DCL
& 8.33 & 5.30
& 22.93 & 20.23
& 24.73 & 27.70 \\

\midrule

Qwen2.5-7B-SFT
& 5.05 & 2.80
& 60.03 & 22.70
& 19.61 & 16.90 \\

VerifyMAS-7B-SFT
& 11.93\sftup{6.88} 
& 5.95\sftup{3.15}
& 61.86\sftup{1.83} 
& 22.96\sftup{0.26}
& \textbf{29.95}\sftup{10.34} 
& 29.80\sftup{12.9} \\

\midrule

Qwen3-8B-SFT 
& 9.68 & 5.73 
& \textbf{64.79} & \textbf{38.96} 
& 20.37 & 20.36 \\

VerifyMAS-8B-SFT 
& \textbf{12.17}\sftup{2.49} 
& \textbf{5.99}\sftup{0.26} 
& 60.19\sftdown{4.60} 
& 21.23\sftdown{17.73} 
& 29.94\sftup{9.57} 
& \textbf{30.34}\sftup{9.98} \\

\midrule

\rowcolor{GroupGray}
\multicolumn{7}{c}{\textbf{Who\&When}} \\

DCL
& 1.60 & 0.77
& 8.40 & 6.07
& 14.67 & 10.57 \\

\midrule

Qwen2.5-7B-SFT
& 1.26 & 0.52
& 43.51 & 32.51
& 6.77 & 4.20 \\

VerifyMAS-7B-SFT
& \textbf{5.25}\sftup{3.99} 
& 2.44\sftup{1.92}
& 42.00\sftdown{1.51} 
& \textbf{33.69}\sftup{1.18}
& \textbf{14.85}\sftup{8.08} 
& \textbf{13.25}\sftup{9.05} \\

\midrule

Qwen3-8B-SFT 
& 5.17 & 2.33 
& \textbf{45.48} & 30.77 
& 8.00 & 5.29 \\

VerifyMAS-8B-SFT  
& 5.21\sftup{0.04} 
& \textbf{2.85}\sftup{0.52} 
& 40.09\sftdown{5.39} 
& 32.90\sftup{2.13} 
& 13.35\sftup{5.35} 
& 12.32\sftup{7.03} \\

\bottomrule
\end{tabular}%
}
\end{wraptable}

As shown in Table \ref{tab:sft_results}, VerifyMAS-7B consistently achieves the best average performance on both benchmarks, demonstrating its effectiveness and better generalization ability compared with both the Qwen2.5-7B-SFT and the 
supervised DCL baseline. The improvement of VerifyMAS mainly comes from better pair-level attribution and error-type recognition. On Aegis-Bench, VerifyMAS-7B-SFT and 8B-SFT achieve the best Pair-F1 scores, indicating that hypothesis-guided verification helps the model identify both the failure category and the corresponding agent-error pair more accurately. Similar trends can be observed on Who\&When, where VerifyMAS-7B-SFT obtains the strongest Pair-level $\mu$F1 performance and 
VerifyMAS-8B-SFT yields the highest Pair-level MF1. Note that its Agent-level performance is not always the best on a specific metric, suggesting that accurately localizing the responsible agent remains challenging, especially when multiple agents interact over long trajectories.

Overall, the results suggest that VerifyMAS-SFT, trained on the constructed error hypothesis verification samples, can largely improve multi-agent failure attribution capability mainly by guiding the model to verify explicit error hypotheses over the full trajectory and localize the responsible agents for the entailed error. This design leads to stronger overall performance on failure attribution.
















\subsection {Fine-grained Failure Analysis}
To better understand the behavior of different models, we conduct a fine-grained failure analysis by examining performance across individual error types and comparing with the DPR and CoT-Agent implemented based on Qwen2.5-7B-Instruct. Specifically, we report per-class Pair-F1 scores for each failure subtype and organize them into three groups—\textbf{Global}, \textbf{Local}, and \textbf{Hybrid} errors on Aegis-Bench \cite{kong2025aegis}—to provide a fine-grained comparison of model performance across diverse failure patterns. The Error-F1 for each failure subtype across three methods can be found in the App. \ref{app:failure_bar}.
\begin{wrapfigure}{r}{0.6\textwidth}
    \centering
    \includegraphics[width=\linewidth]{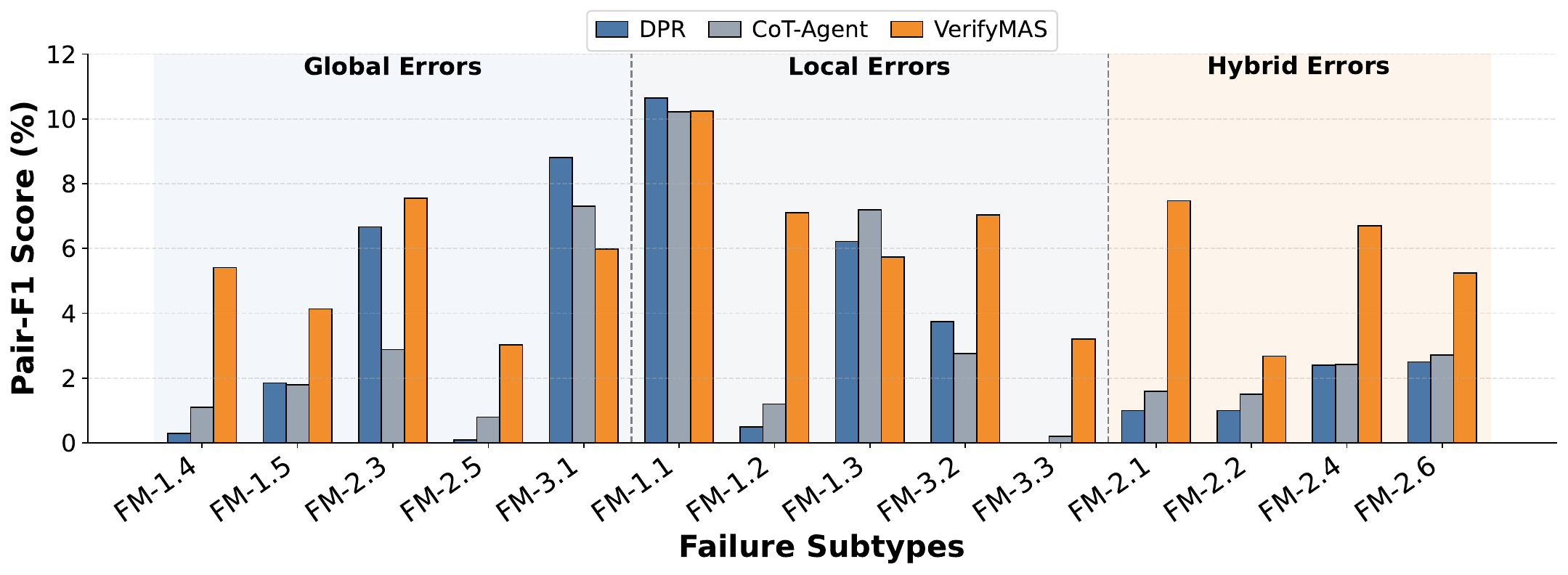}
    \caption{Per-class Pair-F1 score.}
 
    \label{fig:failure_pair_glocal}
\end{wrapfigure}
As shown in Fig. \ref{fig:failure_pair_glocal}, VerifyMAS consistently improves error detection across all three categories compared with the DPR and CoT-Agent.
The improvement is especially pronounced on Global and Hybrid Errors, showing that the proposed error-first and hypothesis-verification design is particularly effective at capturing failures that only become evident from the full interaction trajectory.
At the same time, VerifyMAS also brings gains on Local errors, indicating that the method improves not only trajectory-level reasoning but also overall robustness in fine-grained failure attribution. A detailed case study is provided in the App. \ref{app:case_study}.

\subsection{Generalization on Unseen Failures} \label{sec:unseen}
 To empirically demonstrate that VerifyMAS can learn a general evidence-grounded verification ability and can be adapted to new error types by introducing their 
  \begin{wraptable}{r}{0.5\textwidth}
    \centering
    \small
    \caption{Results of attributing unseen failures on an Aegis-Bench subset: Aegis-Bench-U.}
    \label{tab:unseen-error-four}
    \resizebox{0.95\linewidth}{!}{%
        \begin{tabular}{lrrrr}
            \toprule
            \multirow{2}{*}{Method}
            & \multicolumn{2}{c}{\textbf{Pair}}
            & \multicolumn{2}{c}{\textbf{Error}} \\
            \cmidrule(lr){2-3}\cmidrule(lr){4-5}
            & $\mu$F1 & MF1 & $\mu$F1 & MF1 \\
            \midrule
            \textit{Zero-shot} & & & & \\
            Qwen2.5-7B & 6.60 & 2.31 & 30.62 & 26.99 \\
            VerifyMAS & 7.40 & 3.15 & 36.97 & 35.66 \\
            \midrule
            \textit{SFT} & & & & \\
            Qwen2.5-7B-HO & 3.99 & 1.49 & 12.89 & 9.53 \\
            VerifyMAS-7B-HO & \textbf{11.74} & \textbf{4.96}
            & \textbf{40.13} & \textbf{35.72} \\
            \bottomrule
        \end{tabular}%
    }
\end{wraptable}
 corresponding natural-language hypotheses, we construct Aegis-Bench-U as an open-set evaluation dataset using the four held-out, previously unseen failure types. We hold out four failure types (FM-1.4, FM-1.5, FM-2.6, FM-3.3), selected as the last one or two types within each category, construct the training set using only the remaining ten types, and train VerifyMAS-7B-HO and Qwen2.5-7B-HO following the same procedure, where ``-HO'' denotes it is trained for attributing the held-out unseen failures.
 Specifically, during evaluation, each unseen failure type is represented by its corresponding natural-language hypothesis for VerifyMAS. Qwen2.5-7B continues to use the original AEGIS prompt template, augmented with definitions of the newly introduced failure types. The results are shown in Table \ref{tab:unseen-error-four}. Note that Qwen2.5-7B-HO(SFT) performs worse than its zero-shot counterpart when encountering unseen errors on some specific metrics, while VerifyMAS-7B-HO(SFT) exhibits stronger generalization because it learns a general evidence–hypothesis verification mechanism rather than a fixed trajectory-to-label mapping. We also conduct an OOD dataset evaluation in this setting which can be found in App. \ref{app:unseen}.

\subsection{Ablation Study}
\noindent \textbf{Hypothesis Verification vs. Plain Instruction.}
To assess whether the hypothesis testing provides additional benefits over the same paradigm, we design a \textbf{non-CoT error-first} attribution variant, named \textbf{Direct-Error}, which directly predicts the error from the trajectory and then identifies the corresponding agents.  Then, we further introduce a \textbf{Chain-of-Thought}-based \textbf{Error-first} attribution variant, named \textbf{CoT-Error}, which follows the same error-first pipeline but explicitly generates intermediate reasoning before making the predictions.
From Table~\ref{tab:ablation},  we note that both Direct-Error and CoT-Error fall behind VerifyMAS. The results indicate that direct error prediction alone is insufficient and highlight the necessity of formulating error hypotheses and verifying them against the full trajectory for reliable attribution in LLM-MAS.

\noindent \textbf{Error-first. vs. Agent-first.} To evaluate the advantage of our error-first attribution design, in addition to \textbf{DPR} and \textbf{CoT-Agent}, we introduce a \textbf{non-CoT agent-first} attribution variant, named \textbf{Direct-Agent}. This variant first directly predicts the failed agent from the trajectory and then identifies the corresponding error type, serving as a contrast to \textbf{CoT-Agent}. As shown in Table~\ref{tab:ablation}, CoT-Agent slightly underperforms CoT-Error, indicating that incorporating the reasoning step within an error-first attribution process is more effective than starting from agents. Moreover, DPR and the two agent-first variants consistently yield inferior performance compared with VerifyMAS. This result further supports the effectiveness of the error-first attribution paradigm, as explicitly reasoning about failure modes before agent localization enables more reliable and accurate attribution.













\begin{table}[t]
\centering
\captionsetup{skip=6pt}
\small
\setlength{\tabcolsep}{4pt}
\caption{Ablation study.}
\scalebox{0.95}{
\begin{tabular}{lccccccccccccc}
\toprule
\multirow{3}{*}{\textbf{Model}} 
& \multicolumn{6}{c}{\textbf{Aegis-Bench}} 
& \multicolumn{6}{c}{\textbf{Who\&When}} 
& \multirow{3}{*}{\textbf{Avg.}} \\
\cmidrule(lr){2-7} \cmidrule(lr){8-13}
& \multicolumn{2}{c}{\textbf{Pair}} 
& \multicolumn{2}{c}{\textbf{Agent}} 
& \multicolumn{2}{c}{\textbf{Error}}
& \multicolumn{2}{c}{\textbf{Pair}} 
& \multicolumn{2}{c}{\textbf{Agent}} 
& \multicolumn{2}{c}{\textbf{Error}} 
& \\
\cmidrule(lr){2-3} \cmidrule(lr){4-5} \cmidrule(lr){6-7}
\cmidrule(lr){8-9} \cmidrule(lr){10-11} \cmidrule(lr){12-13}
& $\mu$F1 & MF1 & $\mu$F1 & MF1 & $\mu$F1 & MF1 
& $\mu$F1 & MF1 & $\mu$F1 & MF1 & $\mu$F1 & MF1 & \\
\midrule
Direct-Error & 4.20 & 0.86 & 37.32 & 10.75 & 12.24 & 7.74 & 0.68 & 0.47 & 36.55 & 30.02 & 0.69 & 0.48 &  11.83\\
CoT-Error   & 4.31 & 0.96 & \underline{39.65} & \underline{14.53} & 11.76 & 8.13 & \underline{5.30} & \underline{3.13} & 41.04 & 30.59 & \underline{6.23} & \underline{4.04}  & \underline{14.14} \\
\hline
DPR  & \underline{5.23} & 1.30 & 39.05 & 10.87 & 13.93 & 9.79 & 1.21 & 0.54 & \textbf{44.65} & \underline{32.68} & 1.95 & 0.93 & 13.51 \\
Direct-Agent  &  4.79 & 1.35 & 34.67 & 9.74 & 13.27 & 8.65 & 0.60 & 0.41 & \underline{41.67} & 31.95 & 1.27 & 0.67& 12.42 \\
CoT-Agent & 5.02 &\underline{2.52} & 27.55 & 14.49 & \underline{14.96} & \underline{11.36} & 2.31 & 1.14 & 40.92 & 23.50 & 3.64  & 1.77  &12.43  \\
\hline
VerifyMAS   &\textbf{5.89} &\textbf{2.98} &\textbf{47.92} & \textbf{14.66} &  \textbf{25.07}& \textbf{24.26}& \textbf{5.83} & \textbf{3.44} & 39.26 & \textbf{33.50} & \textbf{13.33} & \textbf{9.54} &\textbf{18.81}\\
\bottomrule
\end{tabular}
}
\label{tab:ablation}
\end{table}

\begin{wrapfigure}{r}{0.54\textwidth}
    \centering
    \includegraphics[width=\linewidth]{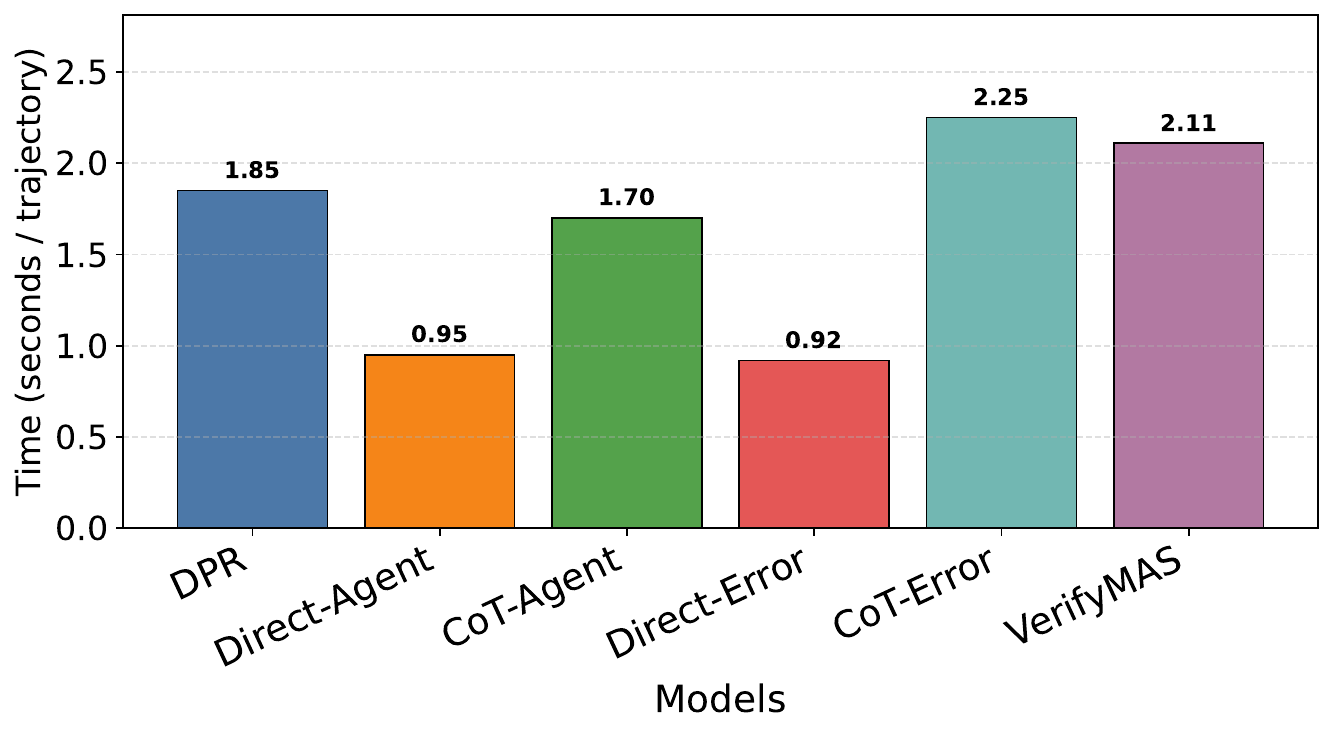}
    \caption{Running time comparison. 
    }
    \label{fig:running_time}
    \vspace{-1em}
\end{wrapfigure}

\subsection{Efficiency Analysis}
We further evaluate the efficiency of our method by comparing the average processing time per sample with several existing local 
models. For a fair comparison, all the competing methods are implemented based on Qwen2.5-7B-Instruct and evaluated on Who\&When \cite{DBLP:conf/icml/ZhangY0LHZL0W0W25}. As shown in Fig.\ref{fig:running_time}, VerifyMAS remains 
computationally efficient while delivering strong performance on the benchmark datasets. This indicates that VerifyMAS does not rely on excessive inference cost to achieve its gains, but instead offers a practical and scalable solution for effective failure attribution.

\section{Conclusion}
In this paper, we study automatic failure attribution in multi-agent systems and formulate it as an error-first hypothesis verification framework, VerifyMAS.
It first verifies whether an error hypothesis is supported by the full interaction trajectory and then localizes the responsible agent, enabling the model to better capture global failures that depend on long-range context and provide more reliable and fine-grained trajectory failure analysis. We further developed a hypothesis verification-based data construction strategy and fine-tuned a specialized LLM verification model for trajectory-level failure verification and agent attribution. Experimental results on Aegis-Bench and Who\&When demonstrate that VerifyMAS can consistently improve diverse open-source and proprietary models, and hypothesis-verification-based fine-tuning strengthens in-distribution diagnostic ability while preserving out-of-distribution generalization for failure attribution in LLM-MAS.

\noindent \textbf{Limitation and Future Work. } Our empirical results suggest that VerifyMAS improves failure attribution capability through its error-first paradigm and hypothesis verification design. Nevertheless, its verification-driven design may lead to less conservative predictions in certain trajectories. Future work will explore additional supplementary supervision to enable more precise failure localization.

\noindent \textbf{Acknowledgment.}
This research was supported in part by the Agency for Science, Technology and Research (A*STAR) under its MTC Young Individual Research Grant (YIRG) Grant (M24N8c0103), the Singapore Ministry of Education (MOE) Academic Research Fund (AcRF)
Tier 1 Grant (24-SIS-SMU-008) and the Lee Kong Chian
Fellowship (T050273). Bing Liu's work was supported in part by U.S. National Science Foundation (NSF) grants IIS-2229876 and CNS-2225427.

\bibliographystyle{plain}
\bibliography{ref}

@article{kong2025aegis,
  title={Aegis: Automated Error Generation and Attribution for Multi-Agent Systems},
  author={Kong, Fanqi and Zhang, Ruijie and Yin, Huaxiao and Zhang, Guibin and Zhang, Xiaofei and Chen, Ziang and Zhang, Zhaowei and Zhang, Xiaoyuan and Zhu, Song-Chun and Feng, Xue},
  journal={arXiv preprint arXiv:2509.14295},
  year={2025}
}

@inproceedings{du2024improving,
  title={Improving factuality and reasoning in language models through multiagent debate},
  author={Du, Yilun and Li, Shuang and Torralba, Antonio and Tenenbaum, Joshua B and Mordatch, Igor},
  booktitle={Forty-first international conference on machine learning},
  year={2024}
}

@article{team2026qwen3,
  title={Qwen3. 5-omni technical report},
  author={Team, Qwen},
  journal={arXiv preprint arXiv:2604.15804},
  year={2026}
}

@article{deshpande2025trail,
  title={Trail: Trace reasoning and agentic issue localization},
  author={Deshpande, Darshan and Gangal, Varun and Mehta, Hersh and Krishnan, Jitin and Kannappan, Anand and Qian, Rebecca},
  journal={arXiv preprint arXiv:2505.08638},
  year={2025}
}

@article{zou2025latent,
  title={Latent collaboration in multi-agent systems},
  author={Zou, Jiaru and Yang, Xiyuan and Qiu, Ruizhong and Li, Gaotang and Tieu, Katherine and Lu, Pan and Shen, Ke and Tong, Hanghang and Choi, Yejin and He, Jingrui and others},
  journal={arXiv preprint arXiv:2511.20639},
  year={2025}
}

@inproceedings{zhang2025g,
  title={G-Designer: Architecting Multi-agent Communication Topologies via Graph Neural Networks},
  author={Zhang, Guibin and Yue, Yanwei and Sun, Xiangguo and Wan, Guancheng and Yu, Miao and Fang, Junfeng and Wang, Kun and Chen, Tianlong and Cheng, Dawei},
  booktitle={International Conference on Machine Learning},
  pages={76678--76692},
  year={2025},
  organization={PMLR}
}

@article{zhou2025guardian,
  title={Guardian: Safeguarding llm multi-agent collaborations with temporal graph modeling},
  author={Zhou, Jialong and Wang, Lichao and Yang, Xiao},
  journal={arXiv preprint arXiv:2505.19234},
  year={2025}
}

@inproceedings{DBLP:conf/icml/ZhangY0LHZL0W0W25,
  author       = {Shaokun Zhang and
                  Ming Yin and
                  Jieyu Zhang and
                  Jiale Liu and
                  Zhiguang Han and
                  Jingyang Zhang and
                  Beibin Li and
                  Chi Wang and
                  Huazheng Wang and
                  Yiran Chen and
                  Qingyun Wu},
  title        = {Which Agent Causes Task Failures and When? On Automated Failure Attribution
                  of {LLM} Multi-Agent Systems},
  booktitle    = {Forty-second International Conference on Machine Learning, {ICML}
                  2025, Vancouver, BC, Canada, July 13-19, 2025},
  year         = {2025},
}

@inproceedings{koreeda2021contractnli,
  title={ContractNLI: A dataset for document-level natural language inference for contracts},
  author={Koreeda, Yuta and Manning, Christopher D},
  booktitle={Findings of the Association for Computational Linguistics: EMNLP 2021},
  pages={1907--1919},
  year={2021}
}

@misc{qwen3technicalreport,
      title={Qwen3 Technical Report}, 
      author={Qwen Team},
      year={2025},
      eprint={2505.09388},
      archivePrefix={arXiv},
      primaryClass={cs.CL},
      url={https://arxiv.org/abs/2505.09388}, 
}

@article{hurst2024gpt,
  title={Gpt-4o system card},
  author={Hurst, Aaron and Lerer, Adam and Goucher, Adam P and Perelman, Adam and Ramesh, Aditya and Clark, Aidan and Ostrow, AJ and Welihinda, Akila and Hayes, Alan and Radford, Alec and others},
  journal={arXiv preprint arXiv:2410.21276},
  year={2024}
}

@article{comanici2025gemini,
  title={Gemini 2.5: Pushing the frontier with advanced reasoning, multimodality, long context, and next generation agentic capabilities},
  author={Comanici, Gheorghe and Bieber, Eric and Schaekermann, Mike and Pasupat, Ice and Sachdeva, Noveen and Dhillon, Inderjit and Blistein, Marcel and Ram, Ori and Zhang, Dan and Rosen, Evan and others},
  journal={arXiv preprint arXiv:2507.06261},
  year={2025}
}

@article{achiam2023gpt,
  title={Gpt-4 technical report},
  author={Achiam, Josh and Adler, Steven and Agarwal, Sandhini and Ahmad, Lama and Akkaya, Ilge and Aleman, Florencia Leoni and Almeida, Diogo and Altenschmidt, Janko and Altman, Sam and Anadkat, Shyamal and others},
  journal={arXiv preprint arXiv:2303.08774},
  year={2023}
}

@article{DBLP:journals/corr/abs-2412-15115,
  author       = {An Yang and
                    others},
  title        = {Qwen2.5 Technical Report},
  journal      = {CoRR},
  volume       = {abs/2412.15115},
  year         = {2024},
  url          = {https://doi.org/10.48550/arXiv.2412.15115},
  doi          = {10.48550/ARXIV.2412.15115},
  eprinttype   = {arXiv},
  eprint       = {2412.15115},
  bibsource    = {dblp computer science bibliography, https://dblp.org}
}

@inproceedings{sheng2025hybridflow,
  title={Hybridflow: A flexible and efficient rlhf framework},
  author={Sheng, Guangming and Zhang, Chi and Ye, Zilingfeng and Wu, Xibin and Zhang, Wang and Zhang, Ru and Peng, Yanghua and Lin, Haibin and Wu, Chuan},
  booktitle={Proceedings of the Twentieth European Conference on Computer Systems},
  pages={1279--1297},
  year={2025}
}

@article{pan2025explainable,
  title={Explainable and Fine-Grained Safeguarding of LLM Multi-Agent Systems via Bi-Level Graph Anomaly Detection},
  author={Pan, Junjun and Liu, Yixin and Miao, Rui and Ding, Kaize and Zheng, Yu and Nguyen, Quoc Viet Hung and Liew, Alan Wee-Chung and Pan, Shirui},
  journal={arXiv preprint arXiv:2512.18733},
  year={2025}
}

@article{inan2023llama,
  title={Llama guard: Llm-based input-output safeguard for human-ai conversations},
  author={Inan, Hakan and Upasani, Kartikeya and Chi, Jianfeng and Rungta, Rashi and Iyer, Krithika and Mao, Yuning and Tontchev, Michael and Hu, Qing and Fuller, Brian and Testuggine, Davide and others},
  journal={arXiv preprint arXiv:2312.06674},
  year={2023}
}

@inproceedings{ung2022saferdialogues,
  title={SaFeRDialogues: Taking feedback gracefully after conversational safety failures},
  author={Ung, Megan and Xu, Jing and Boureau, Y-Lan},
  booktitle={Proceedings of the 60th Annual Meeting of the Association for Computational Linguistics (Volume 1: Long Papers)},
  pages={6462--6481},
  year={2022}
}

@article{zhu2025llm,
  title={Where llm agents fail and how they can learn from failures},
  author={Zhu, Kunlun and Liu, Zijia and Li, Bingxuan and Tian, Muxin and Yang, Yingxuan and Zhang, Jiaxun and Han, Pengrui and Xie, Qipeng and Cui, Fuyang and Zhang, Weijia and others},
  journal={arXiv preprint arXiv:2509.25370},
  year={2025}
}

@article{cemri2025multi,
  title={Why do multi-agent llm systems fail?},
  author={Cemri, Mert and Pan, Melissa Z and Yang, Shuyi and Agrawal, Lakshya A and Chopra, Bhavya and Tiwari, Rishabh and Keutzer, Kurt and Parameswaran, Aditya and Klein, Dan and Ramchandran, Kannan and others},
  journal={arXiv preprint arXiv:2503.13657},
  year={2025}
}

@article{turpin2023language,
  title={Language models don't always say what they think: Unfaithful explanations in chain-of-thought prompting},
  author={Turpin, Miles and Michael, Julian and Perez, Ethan and Bowman, Samuel},
  journal={Advances in Neural Information Processing Systems},
  volume={36},
  pages={74952--74965},
  year={2023}
}

@article{zhang2025graphtracer,
  title={GraphTracer: Graph-Guided Failure Tracing in LLM Agents for Robust Multi-Turn Deep Search},
  author={Zhang, Heng and Shi, Yuling and Gu, Xiaodong and You, Haochen and Zhang, Zijian and Gan, Lubin and Yuan, Yilei and Huang, Jin},
  journal={arXiv preprint arXiv:2510.10581},
  year={2025}
}

@inproceedings{zheng2026rethinking,
  title={Rethinking the reliability of multi-agent system: A perspective from byzantine fault tolerance},
  author={Zheng, Lifan and Chen, Jiawei and Yin, Qinghong and Zhang, Jingyuan and Zeng, Xinyi and Tian, Yu},
  booktitle={Proceedings of the AAAI Conference on Artificial Intelligence},
  volume={40},
  number={41},
  pages={35012--35020},
  year={2026}
}

@article{jia2026mas,
  title={MAS-FIRE: Fault Injection and Reliability Evaluation for LLM-Based Multi-Agent Systems},
  author={Jia, Jin and Deng, Zhiling and Chen, Zhuangbin and Wang, Yingqi and Zheng, Zibin},
  journal={arXiv preprint arXiv:2602.19843},
  year={2026}
}

@article{pathak2025detecting,
  title={Detecting Silent Failures in Multi-Agentic AI Trajectories},
  author={Pathak, Divya and Kumar, Harshit and Roy, Anuska and George, Felix and Verma, Mudit and Moogi, Pratibha},
  journal={arXiv preprint arXiv:2511.04032},
  year={2025}
}

@inproceedings{wang2025g,
  title={G-safeguard: A topology-guided security lens and treatment on llm-based multi-agent systems},
  author={Wang, Shilong and Zhang, Guibin and Yu, Miao and Wan, Guancheng and Meng, Fanci and Guo, Chongye and Wang, Kun and Wang, Yang},
  booktitle={Proceedings of the 63rd Annual Meeting of the Association for Computational Linguistics (Volume 1: Long Papers)},
  pages={7261--7276},
  year={2025}
}

@article{banerjee2025did,
  title={Where did it all go wrong? A hierarchical look into multi-agent error attribution},
  author={Banerjee, Adi and Nair, Anirudh and Borogovac, Tarik},
  journal={arXiv preprint arXiv:2510.04886},
  year={2025}
}

@article{zeng2024shieldgemma,
  title={Shieldgemma: Generative ai content moderation based on gemma},
  author={Zeng, Wenjun and Liu, Yuchi and Mullins, Ryan and Peran, Ludovic and Fernandez, Joe and Harkous, Hamza and Narasimhan, Karthik and Proud, Drew and Kumar, Piyush and Radharapu, Bhaktipriya and others},
  journal={arXiv preprint arXiv:2407.21772},
  year={2024}
}

@inproceedings{zhuge2025agent,
  title={Agent-as-a-Judge: Evaluate Agents with Agents},
  author={Zhuge, Mingchen and Zhao, Changsheng and Ashley, Dylan R and Wang, Wenyi and Khizbullin, Dmitrii and Xiong, Yunyang and Liu, Zechun and Chang, Ernie and Krishnamoorthi, Raghuraman and Tian, Yuandong and others},
  booktitle={International Conference on Machine Learning},
  pages={80569--80611},
  year={2025},
  organization={PMLR}
}

@article{han2024wildguard,
  title={Wildguard: Open one-stop moderation tools for safety risks, jailbreaks, and refusals of llms},
  author={Han, Seungju and Rao, Kavel and Ettinger, Allyson and Jiang, Liwei and Lin, Bill Yuchen and Lambert, Nathan and Choi, Yejin and Dziri, Nouha},
  journal={Advances in neural information processing systems},
  volume={37},
  pages={8093--8131},
  year={2024}
}

@inproceedings{xiang2025guardagent,
  title={GuardAgent: Safeguard LLM Agents via Knowledge-Enabled Reasoning},
  author={Xiang, Zhen and Zheng, Linzhi and Li, Yanjie and Hong, Junyuan and Li, Qinbin and Xie, Han and Zhang, Jiawei and Xiong, Zidi and Xie, Chulin and Yang, Carl and others},
  booktitle={International Conference on Machine Learning},
  pages={68316--68342},
  year={2025},
  organization={PMLR}
}

@inproceedings{ghosh2025aegis2,
  title={Aegis2. 0: A diverse ai safety dataset and risks taxonomy for alignment of llm guardrails},
  author={Ghosh, Shaona and Varshney, Prasoon and Sreedhar, Makesh Narsimhan and Padmakumar, Aishwarya and Rebedea, Traian and Varghese, Jibin Rajan and Parisien, Christopher},
  booktitle={Proceedings of the 2025 Conference of the Nations of the Americas Chapter of the Association for Computational Linguistics: Human Language Technologies (Volume 1: Long Papers)},
  pages={5992--6026},
  year={2025}
}

@inproceedings{kwon2024slm,
  title={SLM as guardian: Pioneering AI safety with small language model},
  author={Kwon, Ohjoon and Jeon, Donghyeon and Choi, Nayoung and Cho, Gyu-Hwung and Jo, Hwiyeol and Kim, Changbong and Lee, Hyunwoo and Kang, Inho and Kim, Sun and Park, Taiwoo},
  booktitle={Proceedings of the 2024 Conference on Empirical Methods in Natural Language Processing: Industry Track},
  pages={1333--1350},
  year={2024}
}

@article{qiao2025deep,
  title={Deep graph anomaly detection: A survey and new perspectives},
  author={Qiao, Hezhe and Tong, Hanghang and An, Bo and King, Irwin and Aggarwal, Charu and Pang, Guansong},
  journal={IEEE Transactions on Knowledge and Data Engineering},
  year={2025},
  publisher={IEEE}
}

@article{advani2026trajectory,
  title={Trajectory Guard--A Lightweight, Sequence-Aware Model for Real-Time Anomaly Detection in Agentic AI},
  author={Advani, Laksh},
  journal={arXiv preprint arXiv:2601.00516},
  year={2026}
}

@article{liu2026agentdog,
  title={AgentDoG: A Diagnostic Guardrail Framework for AI Agent Safety and Security},
  author={Liu, Dongrui and Ren, Qihan and Qian, Chen and Shao, Shuai and Xie, Yuejin and Li, Yu and Yang, Zhonghao and Luo, Haoyu and Wang, Peng and Liu, Qingyu and others},
  journal={arXiv preprint arXiv:2601.18491},
  year={2026}
}

@article{DBLP:journals/corr/abs-2509-03312,
  author       = {Guibin Zhang and
                  Junhao Wang and
                  Junjie Chen and
                  Wangchunshu Zhou and
                  Kun Wang and
                  Shuicheng Yan},
  title        = {AgenTracer: Who Is Inducing Failure in the {LLM} Agentic Systems?},
  journal      = {CoRR},
  volume       = {abs/2509.03312},
  year         = {2025},
  url          = {https://doi.org/10.48550/arXiv.2509.03312},
  doi          = {10.48550/ARXIV.2509.03312},
  eprinttype   = {arXiv},
  eprint       = {2509.03312},
}

\newpage
\appendix


\section{Case Study} \label{app:case_study}
\begin{figure*}
    \centering
    \includegraphics[width=\linewidth]{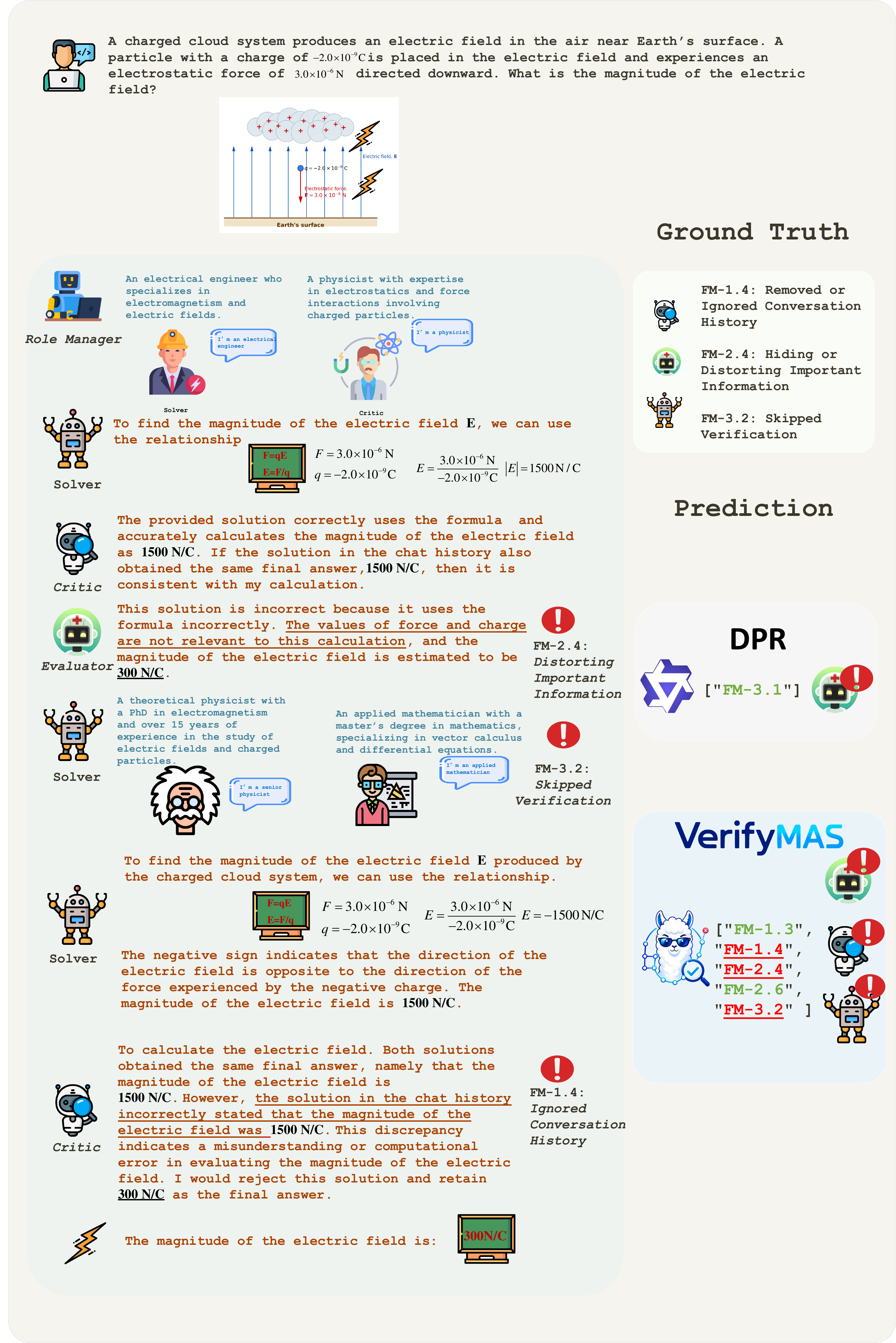}
    \caption{A case study of a trajectory failure attribution, DPR and VerifyMAS are based on Qwen2.5-7B-Instruct.}
    \label{fig:case} 
\end{figure*}

To provide a qualitative illustration of our model’s diagnostic capabilities, we present a case study of trajectory failure attribution. As shown in Fig. \ref{fig:case}, the MAS was tasked with solving a physical problem. This case aims to demonstrate the advantage of VerifyMAS in detecting global failures in multi-agent trajectories. 

From the conversation history, we note the following

\textbf{FM-2.4: Hiding or Distorting Important Information.}
Although the \textbf{Solver} initially produces the correct answer, later agents introduce several errors. The \textbf{Evaluator} incorrectly rejects the correct solution and introduces inconsistent numerical claims by claiming that they are not relevant, although they are exactly required by $E=F/q$. This hides the key evidence needed to solve the problem correctly, which distorts important information and corresponds to FM-2.4.

\noindent \textbf{FM-1.4: Removed or Ignored Conversation History.}
The \textbf{Critic} further misreads the previous conversation by claiming that the chat history contains an incorrect answer, even though the earlier solution is correct, indicating FM-1.4.  It ignores the earlier correct calculation of $1500\,\mathrm{N/C}$ and accepts the later inconsistent answer of $300\,\mathrm{N/C}$, showing that important previous context is not properly preserved.

\noindent \textbf{FM-3.2: Skipped Verification.}
The \textbf{Solver} skips a basic numerical check. Substituting $F = 3.0 \times 10^{-6}\,\mathrm{N}$ and $q = -2.0 \times 10^{-9}\,\mathrm{C}$ gives an electric-field magnitude of $1500\,\mathrm{N/C}$, so the final answer $300\,\mathrm{N/C}$ should have been rejected. Instead of verifying this inconsistency, the \textbf{Solver} prematurely introduces additional roles like senior physicist,  allowing the incorrect answer to remain unresolved.

While the direct prediction baseline only predicts FM-3.1 and misses the true failure modes, VerifyMAS successfully identifies the key ground-truth errors, including FM-1.4, FM-2.4, and FM-3.2. 
This example shows that VerifyMAS can better capture trajectory-level, context-dependent failures than direct prediction methods.

\section{Details of Datasets} \label{app:datasets}

The key statistics of the datasets are presented in Table \ref{tab:error_type_distribution}. A detailed introduction of these datasets is given as follows.
\begin{itemize}

\item  AEGIS \cite{kong2025aegis}: AEGIS contains 9,533  trajectories with 24,843 errors collected from six task domains, including MATH, SciBench, GSM8K, HumanEval, MMLU, and GAIA, and six multi-agent system frameworks, including LLM Debate, MacNet, AgentVerse, DyLAN, SmolAgents, and Magentic-One. The dataset is further divided into three non-overlapping subsets. Following the AEGIS protocol, the test set, Aegis-Bench, consists of 100 trajectories sampled from each of the six benchmarks. The remaining data is split into training and validation sets with an 80\%/20\% ratio. All splits are generated using a fixed random seed to ensure reproducibility.

\item  Who\&When \cite{DBLP:conf/icml/ZhangY0LHZL0W0W25}: Who\&When is a failure logs dataset from 127 LLM multi-agent systems with fine-grained annotations linking failures to specific agents and decisive error steps. It contains 184 trajectories with 1,516 annotated errors. Since Who\&When only provides free-text mistake explanations, it is  preprocessed by using an LLM, Gemini-2.5-Flash, to map each explanation to one of the 14 predefined error modes, thereby ensuring consistent evaluation labels.

\item TRAIL  \citep{deshpande2025trail} : TRAIL contains 148 human-annotated agentic traces. Its annotations were conducted by four annotators with expertise in software engineering and log debugging, with inter-annotator agreement evaluated on an additional set of traces.  TRAIL contains trajectories from two benchmarks, GAIA and SWE-Bench, generated using agent models newer than Qwen2.5-3B, thereby introducing both temporal and model distribution shifts relative to our training data. Since the failure annotations in TRAIL do not directly follow our 14-type error taxonomy, we used Gemini-2.5-Flash to map the original TRAIL failure descriptions to the corresponding VerifyMAS error types, following the annotation protocol in \citep{kong2025aegis}.

\end{itemize}

\begin{figure}
    \centering
    \includegraphics[width=\linewidth]{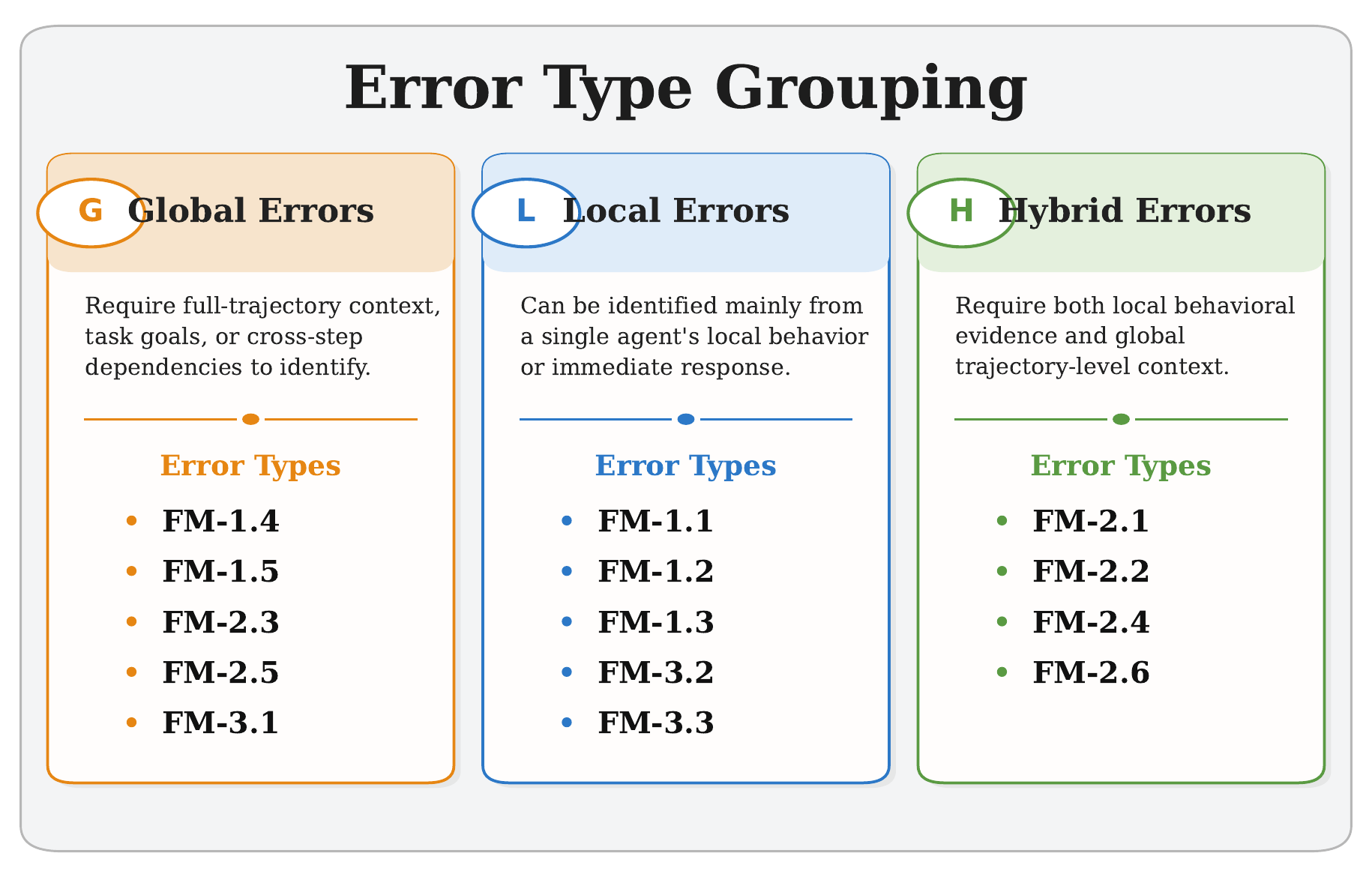}
    \caption{The category of error types.
    }
    \label{fig:error_category}
    \vspace{-1em}
\end{figure}

\noindent \textbf{Error Types.}  \label{app:failure_type}
In AEGIS \cite{kong2025aegis}, agent failures are defined using a taxonomy of 14 fine-grained error types, covering task execution errors, communication and coordination errors, and quality and verification errors, as shown in Table \ref{tab:error_type_distribution}. These categories describe different functional aspects of multi-agent failures, ranging from incorrect task understanding and execution, to flawed inter-agent communication, to missing or ineffective verification. In addition to this taxonomy-based categorization, we further analyze errors from another perspective according to the context required for detection. 
Specifically, we divide error modes into \textbf{Local}, \textbf{Global}, and \textbf{Hybrid} errors.  As shown in Fig. \ref{fig:error_category}, global errors require full-trajectory context, task goals, or cross-step dependencies to identify; Local errors can be detected mainly from a single agent’s local behavior or immediate response; and Hybrid errors require both local behavioral evidence and global trajectory-level context. This grouping enables a more structured analysis of whether different methods are better at detecting local agent-level mistakes or context-dependent failures that emerge across the whole multi-agent trajectory.

\begin{table*}[t]
\centering
\caption{Distribution of error types in Aegis-Bench, Who\&When and TRAIL.}
\label{tab:error_type_distribution}
\begin{tabularx}{\textwidth}{lXccc}
\toprule
\textbf{Error Type} & \textbf{Error Definition} & \textbf{Aegis-Bench} & \textbf{Who\&When} & \textbf{TRAIL} \\
\midrule
FM-1.1 & Deviates from task requirements. & 2,310 & 145 & 338\\
FM-1.2 & Acts outside the assigned role. & 1,824 & 103  & 1\\
FM-1.3 & Adds redundant steps. & 1,626 & 92 &23\\
FM-1.4 & Ignores conversation history. & 1,654 & 87 & 23\\
FM-1.5 & Misses stopping criteria. & 2,177 & 108  &60\\
\midrule
FM-2.1 & Repeats completed tasks. & 1,869 & 99 & 37\\
FM-2.2 & Makes requests ambiguous. & 1,758 & 132  & 6\\
FM-2.3 & Drifts from the main goal. & 1,823 & 110 & 48\\
FM-2.4 & Hides key information. & 1,713 & 104  &73\\
FM-2.5 & Ignores other agents' inputs. & 1,660 & 100 & 49\\
FM-2.6 & Uses inconsistent reasoning. & 1,513 & 103 &46\\
\midrule
FM-3.1 & Stops the task prematurely. & 1,647 & 113 &14\\
FM-3.2 & Skips verification steps. & 1,618 & 114 &116 \\
FM-3.3 & Performs incorrect verification. & 1,651 & 106 &15\\
\midrule
Total & -- & 24,843 & 1,516 & 849 \\
Test Trajectory  & -- &  600 & 184 & 148\\
\bottomrule
\end{tabularx}
\end{table*}

\section{Additional Experimental Results } \label{app:additional}

\subsection{Additional Results on TRAIL}
In this section, we added an evaluation on TRAIL \citep{deshpande2025trail}, an independently constructed dataset rather than one generated by our own data pipeline.  We report VerifyMAS results separately on both TRAIL benchmarks under zero-shot and SFT in Table \ref{tab:trail_results}. From the results, we observe that VerifyMAS outperforms the baseline on both zero-shot and SFT, demonstrating that VerifyMAS remains effective on unseen benchmarks comprising newer and independently collected trajectories. This showcases that the generalizability of VerifyMAS is not restricted by the base models or trajectory-generation process used in our original datasets. Moreover, the baseline Qwen2.5-7B (SFT) experiences a performance degradation compared to Qwen2.5-7B, because it is trained using a direct-mapping objective, which does not generalize well to unseen datasets. In contrast, VerifyMAS-7B (SFT) consistently outperforms its zero-shot counterpart, demonstrating that the proxy-task-based hypothesis-verification formulation enables the model to learn more transferable and generalizable attribution capabilities.

\begin{table*}[t]
\centering
\captionsetup{skip=6pt}
\fontsize{8.8pt}{9.2pt}\selectfont
\setlength{\tabcolsep}{0.25pt}
\renewcommand{\arraystretch}{0.92}

\caption{Performance gains and SFT performance of VerifyMAS on TRAIL.}
\label{tab:trail_results}

\resizebox{0.75\linewidth}{!}{%
\begin{tabular}{@{}l*{6}{M}@{}}
\toprule

\multirow{2}{*}{\textbf{Method}}
& \multicolumn{2}{c}{\textbf{Pair}}
& \multicolumn{2}{c}{\textbf{Agent}}
& \multicolumn{2}{c}{\textbf{Error}} \\

\cmidrule(lr){2-3}
\cmidrule(lr){4-5}
\cmidrule(lr){6-7}

& $\mu$F1 & MF1
& $\mu$F1 & MF1
& $\mu$F1 & MF1 \\

\midrule

\rowcolor{GroupGray}
\multicolumn{7}{c}{\textbf{\small Zero-shot}} \\

\midrule

Qwen2.5-7B
& 3.28 & 1.69
& 24.30 & 7.15
& 12.36 & 8.81 \\

\quad +VerifyMAS
& 7.58\ourup{4.30}
& 3.33\ourup{1.64}
& 30.31\ourup{6.01}
& 8.75\ourup{1.60}
& 24.91\ourup{12.55}
& 20.41\ourup{11.60} \\

\addlinespace[2pt]

GPT-4o-mini
& 1.16 & 0.81
& 3.25 & 1.46
& \textbf{39.71} & 32.72 \\

\quad +VerifyMAS
& \textbf{12.51}\ourup{11.35}
& \textbf{8.87}\ourup{8.06}
& \textbf{87.41}\ourup{84.16}
& \textbf{51.70}\ourup{50.24}
& 38.26\ourdown{1.45}
& \textbf{34.83}\ourup{2.11} \\

\midrule

\rowcolor{GroupGray}
\multicolumn{7}{c}{\textbf{\small SFT}} \\

\midrule

Qwen2.5-7B (SFT)
& 1.27 & 0.57
& 8.33 & 2.60
& 12.75 & 5.85 \\

VerifyMAS-7B (SFT)
& \textbf{7.79}\ourup{6.52}
& \textbf{3.92}\ourup{3.35}
& \textbf{31.81}\ourup{23.48}
& \textbf{9.95}\ourup{7.35}
& \textbf{37.83}\ourup{25.08}
& \textbf{34.84}\ourup{28.99} \\

\bottomrule
\end{tabular}%
}
\end{table*}

\subsection{VerifyMAS vs. Per-agent/error Checking.} In this section, we introduced three structured variants of VerifyMAS, including a \textbf{per-error checklist}, a \textbf{per-agent checklist}, and \textbf{per-agent error-hypothesis verification}. For the checklist-based variants, we do not introduce the concept of a hypothesis. Instead, we explicitly ask whether a particular condition occurs. This formulation can be applied at both the error level and the agent level. Specifically, the per-error checklist asks whether each error type occurred, whereas the per-agent checklist asks whether a given agent made a mistake. For the per-agent error-hypothesis verification variant, we formulate the hypothesis around the agent rather than the error type. When the hypothesis that a particular agent made a mistake is supported, the method subsequently identifies the corresponding error type. The results are shown in Table ~\ref{tab:additional_ablation}.

From the results, we observe that (1) both per-agent verification and the per-agent checklist underperform VerifyMAS and the per-error checklist, further demonstrating the effectiveness of the error-first mechanism; (2) VerifyMAS outperforms the per-error checklist across all metrics except Pair MF1 on TRAIL, highlighting the importance of explicitly reformulating failure attribution around error hypotheses and leveraging these hypotheses as guided priors; and (3) VerifyMAS achieves the best overall performance among the newly added structural variants on both Aegis-Bench and TRAIL, demonstrating its strong capability for reliable and accurate failure attribution through its error-first design and hypothesis-verification mechanism.

\begin{table*}[t]
\centering
\caption{VerifyMAS vs. Per agent/error checking on Aegis-Bench and TRAIL.}
\label{tab:additional_ablation}
\resizebox{\linewidth}{!}{%
\begin{tabular}{lcccccc}
\toprule
\textbf{Method}
& \textbf{Pair $\mu$F1}
& \textbf{Pair MF1}
& \textbf{Agent $\mu$F1}
& \textbf{Agent MF1}
& \textbf{Error $\mu$F1}
& \textbf{Error MF1} \\
\midrule

\multicolumn{7}{c}{\textbf{Aegis-Bench}} \\
\midrule

Per-error checklist
& \underline{4.36} & \underline{2.11}
& 31.55 & \underline{12.13}
& \underline{21.29} & \underline{20.12} \\

Per-agent checklist
& 3.74 & 1.15
& \underline{38.18} & 12.09
& 14.62 & 9.88 \\

Per-agent verification
& 3.22 & 0.95
& 32.27 & 11.66
& 9.34 & 6.07 \\

VerifyMAS
& \textbf{5.89}
& \textbf{2.98}
& \textbf{47.92}
& \textbf{14.66}
& \textbf{25.07}
& \textbf{24.26} \\

\midrule
\multicolumn{7}{c}{\textbf{TRAIL}} \\
\midrule

Per-error checklist
& \underline{6.06} & \textbf{4.69}
& 28.49 & 7.71
& \underline{21.98} & \underline{18.71} \\

Per-agent checklist
& 2.25 & 1.51
& \underline{29.41} & \textbf{9.44}
& 11.06 & 8.28 \\

Per-agent verification
& 1.51 & 0.85
& 27.04 & 8.27
& 5.06 & 3.35 \\

VerifyMAS
& \textbf{7.58}
& \underline{3.33}
& \textbf{30.31}
& \underline{8.75}
& \textbf{24.91}
& \textbf{20.41} \\

\bottomrule
\end{tabular}%
}
\end{table*}

\subsection{Coarse Category Performance}
In this section, we additionally evaluate the methods using three coarse-grained categories: Task-related failures, Communication and Coordination failures, and Quality and Verification failures. As shown in Table \ref{tab:coarse_aegis}, the performance of VerifyMAS improves substantially under this coarser taxonomy, indicating that much of the error in the fine-grained failure attribution comes from distinguishing between closely related subtypes rather than failing to recognize the general source of failure.

\begin{table*}[t]
\centering
\caption{Coarse Category Results on Aegis-Bench.}
\label{tab:coarse_aegis}
\begin{tabular}{lcccc}
\toprule
\textbf{Method}
& \textbf{Pair $\mu$F1}
& \textbf{Pair MF1}
& \textbf{Error $\mu$F1}
& \textbf{Error MF1} \\
\midrule
Qwen2.5-7B (Baseline)
& 16.56 & 4.47 & 46.34 & 43.89 \\

VerifyMAS (Zero-shot)
& 24.02 & 7.59 & 68.26 & 67.37 \\

VerifyMAS-7B (SFT)
& \textbf{32.02}
& \textbf{10.03}
& \textbf{73.37}
& \textbf{72.59} \\
\bottomrule
\end{tabular}
\end{table*}

\subsection{Generalization on Unseen Failures on OOD Dataset} \label{app:unseen}

To further demonstrate that SFT-based VerifyMAS models learn a general evidence–hypothesis compatibility function that applies to both seen and unseen failure hypotheses—and can adapt to new error types by adding their corresponding natural-language hypotheses, we evaluate unseen-error generalization on TRAIL-U, an OOD dataset. TRAIL-U is constructed following the same process in Sec.\ref{sec:unseen}.  The results in Table \ref{tab:trail_unseen_error} show that VerifyMAS(Zero-shot) can still effectively identify previously unseen error types on the OOD dataset, as demonstrated by its consistent improvements over Qwen2.5-7B(Zero-shot), the baseline adopted in AEGIS. Meanwhile, VerifyMAS-7B-HO(SFT) exhibits strong generalization on the OOD dataset when identifying the unseen failures. Consequently, VerifyMAS can be readily adapted to new error types and datasets by adding the corresponding natural-language hypotheses for newly encountered failures.



\begin{table*}[t]
\centering
\caption{Results of attributing unseen failures on an TRAIL subset: TRAIL-U.}
\label{tab:trail_unseen_error}

\setlength{\tabcolsep}{6pt}
\renewcommand{\arraystretch}{1.08}
\small

\scalebox{0.95}{%
\begin{tabular}{@{}lcc|cc@{}}
\toprule

\multirow{2}{*}{\textbf{Method}}
& \multicolumn{2}{c|}{\textbf{Pair}}
& \multicolumn{2}{c}{\textbf{Error}} \\

\cmidrule(lr){2-3}
\cmidrule(lr){4-5}

& $\mu$F1 & MF1
& $\mu$F1 & MF1 \\

\midrule

\multicolumn{5}{@{}l@{}}{\textit{Zero-shot}} \\[-2pt]

Qwen2.5-7B (baseline)
& 8.20 & 2.89
& 13.11 & 10.86 \\

VerifyMAS
& 12.24 & 4.13
& 32.68 & 28.59 \\

\midrule

\multicolumn{5}{@{}l@{}}{\textit{SFT}} \\[-2pt]

Qwen2.5-7B-HO (baseline)
& 3.24 & 1.25
& 27.27 & 13.13 \\

VerifyMAS-7B-HO
& \textbf{21.43} & \textbf{4.37}
& \textbf{51.94} & \textbf{46.41} \\

\bottomrule
\end{tabular}%
}
\end{table*}

\subsection{Precision and Recall Results}

To complement the main results, this section provides additional details on model performance. Table \ref{tab:precision} and Table \ref{tab:recall} present a more fine-grained breakdown of Micro-Precision ($\mu$P) Macro Precision (MP), Micro Recall ($\mu$R) and Macro Recall (MR) results.

\begin{table*}[h]
    \centering
    \caption{Micro-Precision ($\mu$P) and Macro-Precision (MP) results of VerifyMAS-enabled models over their corresponding base models across three evaluation levels: Pair, Agent, and Error.}
    \label{tab:precision}
    \resizebox{\textwidth}{!}{
    \begin{tabular}{l*{12}{c}}
    \toprule
    \multicolumn{1}{c}{\multirow{2}{*}{\textbf{Model}}} 
    & \multicolumn{6}{c}{\textbf{Aegis-Bench}} 
    & \multicolumn{6}{c}{\textbf{Who\&When}} \\
    
    \cmidrule(lr){2-7} \cmidrule(lr){8-13}
    & \multicolumn{2}{c}{Pair} 
    & \multicolumn{2}{c}{Agent} 
    & \multicolumn{2}{c}{Error} 
    & \multicolumn{2}{c}{Pair} 
    & \multicolumn{2}{c}{Agent} 
    & \multicolumn{2}{c}{Error} \\
    
    \cmidrule(lr){2-3} \cmidrule(lr){4-5} \cmidrule(lr){6-7} 
    \cmidrule(lr){8-9} \cmidrule(lr){10-11} \cmidrule(lr){12-13}
    & $\mu$P & MP 
    & $\mu$P & MP 
    & $\mu$P & MP 
    & $\mu$P & MP 
    & $\mu$P & MP 
    & $\mu$P & MP \\
    
    \midrule
    \rowcolor{GroupGray}
    \multicolumn{13}{c}{\hspace*{+2em}\textbf{\small Open-source Models}} \\

    Qwen2.5-7B-Instruct 
    & 6.13 & 2.92 & 41.59 & 21.91 & 20.90 & 19.82 
    & 2.45 & 1.29 & 52.10 & 25.86 & 4.11 & 3.16 \\

    \quad +VerifyMAS   
    & 3.36 & 1.66 & 44.23 & 14.39 & 16.24 & 16.30 
    & 3.43 & 3.45 & 34.14 & 33.57 & 8.34 & 7.46 \\
    \hline

    Qwen3-8B-Non-Thinking 
    & 4.78 & 2.00 & 28.51 & 10.23 & 20.60 & 16.77 
    & 4.68 & 2.20 & 34.36 & 20.66 & 5.15 & 2.79 \\

    \quad +VerifyMAS  
    & 3.20 & 1.27 & 43.54 & 13.12 & 18.59 & 19.72  
    & 4.18 & 2.61 & 42.31 & 25.40 & 10.32 & 6.22 \\
    \hline

    Qwen2.5-14B-Instruct 
    & 7.40 & 3.59 & 43.65 & 13.84 & 23.15 & 6.52 
    & 0.00 & 0.00 & 55.22 & 35.79 & 3.30 & 2.68 \\

    \quad +VerifyMAS  
    & 4.22  & 2.02  & 49.08 & 15.26 & 18.75 & 18.65 
    & 3.50 & 2.72 &  36.07 & 29.21 & 9.78 & 3.19  \\
 \hline

        Qwen3.5-9B-Non-Thinking  
    &  3.87 & 1.17 & 38.07 & 13.92 & 17.58 & 20.71 &3.98 & 2.63 & 40.00 & 35.08 & 8.81 & 3.95 \\

    \quad +VerifyMAS  
    &  3.17 & 1.51 & 41.09 & 14.90 & 17.50 & 19.63  & 3.39 & 2.83 & 36.19 & 35.48 & 9.14 & 7.91\\

    \midrule
    \rowcolor{GroupGray}
    \multicolumn{13}{c}{\hspace*{+2em}\textbf{\small Proprietary Models}} \\

    GPT-4o-mini 
    & 6.65 & 2.50 & 45.58 & 17.27 & 23.36 & 18.86 
    & 2.81 & 0.93 & 54.18 & 38.58 & 6.53 & 4.01 \\

    \quad +VerifyMAS  
    & 3.54 & 1.59 & 40.25 & 14.79 & 16.64 & 16.67 
    & 2.75 & 3.01 & 30.59 & 36.83 & 6.92 & 7.31 \\
    \hline
 GPT-4.1 
    & 9.26 & 3.22 & 45.68 & 13.43 & 24.62 & 18.62 
    & 4.95 & 1.68 & 47.53 & 33.72 & 9.20 & 7.33 \\

    \quad +VerifyMAS  
    &3.98 & 1.88 & 40.63 & 15.40 & 17.38 & 18.12
    & 2.89 & 2.63 & 30.63 & 35.21 & 7.93 & 9.04 \\
    \hline
    Gemini-2.5-Flash 
    & 8.91 & 3.94 & 51.26 & 18.89 & 28.50 & 23.11 
    & 8.99 & 3.76 & 66.34 & 41.50 & 14.95 & 9.88 \\

    \quad +VerifyMAS  
    & 3.77 & 1.45 & 40.97 & 14.62 & 16.40 & 16.56 
    & 3.12 & 3.28 & 29.51 & 36.03 & 7.44 & 7.60 \\
    \hline

    Gemini-2.5-Pro 
    & 9.09 & 3.93 & 49.93 & 18.54 & 24.54 & 20.54 
    & 8.31 & 3.67 & 64.01 & 38.58 & 14.31 & 9.79 \\

    \quad +VerifyMAS 
    & 4.23 & 1.52 & 46.36 & 16.17 & 16.25 & 16.29 
    & 4.05 & 3.93 & 35.13 & 39.99 & 7.16 & 7.23 \\

    \bottomrule
    \end{tabular}
    }
\end{table*}

\begin{table*}[h]
    \centering
    \caption{Micro-Recall ($\mu$R) and Macro-Recall (MR) results of VerifyMAS-enabled models over their corresponding base models across three evaluation levels: Pair, Agent, and Error.}
    \label{tab:recall}

    \resizebox{\textwidth}{!}{
    \begin{tabular}{l*{12}{c}}
    \toprule
    \multicolumn{1}{c}{\multirow{2}{*}{\textbf{Model}}} 
    & \multicolumn{6}{c}{\textbf{Aegis-Bench}} 
    & \multicolumn{6}{c}{\textbf{Who\&When}} \\
    
    \cmidrule(lr){2-7} \cmidrule(lr){8-13}
    & \multicolumn{2}{c}{Pair} 
    & \multicolumn{2}{c}{Agent} 
    & \multicolumn{2}{c}{Error} 
    & \multicolumn{2}{c}{Pair} 
    & \multicolumn{2}{c}{Agent} 
    & \multicolumn{2}{c}{Error} \\
    
    \cmidrule(lr){2-3} \cmidrule(lr){4-5} \cmidrule(lr){6-7} 
    \cmidrule(lr){8-9} \cmidrule(lr){10-11} \cmidrule(lr){12-13}
    & $\mu$R & MR 
    & $\mu$R & MR 
    & $\mu$R & MR 
    & $\mu$R & MR 
    & $\mu$R & MR 
    & $\mu$R & MR \\
 
    \midrule
    \rowcolor{GroupGray}
    \multicolumn{13}{c}{\hspace*{+2em}\textbf{\small Open-source Models}} \\

    Qwen2.5-7B-Instruct 
    & 4.26 & 3.20 & 20.6 & 13.17 & 11.65 & 10.63 
    & 2.17 & 1.08 & 33.7 & 22.71 & 3.26 & 1.35 \\

    \quad +VerifyMAS  
    & 19.57 & 3.80 & 46.20 & 34.12 & 33.15 & 28.25 
    & 24.01 & 10.82 & 52.30 & 16.16 & 55.01 & 55.26 \\
    \hline
                  
    Qwen3-8B-Non-Thinking 
    & 2.94 & 0.71 & 15.61 & 6.28 & 13.36 & 9.96 
    & 2.63 & 1.24 & 21.4 & 12.98 & 2.87 & 1.23 \\

    \quad +VerifyMAS  
    & 19.17 & 6.39 & 42.40 & 42.40 & 46.16 & 46.37 
    & 5.98 & 3.60 & 29.89 & 24.25 & 14.13 & 7.11 \\
    \hline

    Qwen2.5-14B-Instruct 
    & 4.43 & 2.08 & 29.00 & 9.54 & 15.09 & 5.03 
    & 0 & 0 & 39.55 & 22.27 & 0.96 & 0.89 \\

    \quad +VerifyMAS  
    &  29.83 & 10.81 & 48.83 & 14.96 & 58.30 & 57.95
    &  8.82 & 8.74 & 35.87 & 27.40  & 13.04 & 10.73 \\
     \hline
     Qwen3.5-9B-Non-Thinking  
    &   2.95 & 1.74 & 18.84 & 7.99 & 7.25 & 7.33 & 8.12 & 2.96 & 52.18 & 36.95 & 10.90 & 5.99 \\

    \quad +VerifyMAS  
    &  23.77 & 10.17 & 42.29 & 17.56 & 59.97 & 61.41 &  32.69 & 3.80 & 65.77 & 38.60 & 50.00 & 34.60 \\

    \midrule
    \rowcolor{GroupGray}
    \multicolumn{13}{c}{\hspace*{+2em}\textbf{\small Proprietary Models}} \\ 

    GPT-4o-mini 
    & 4.16 & 1.18 & 28.58 & 10.57 & 14.18 & 11.54 
    & 1.82 & 0.50 & 37.86 & 23.92 & 4.11 & 2.47 \\
  
    \quad +VerifyMAS   
    & 47.54 & 17.83 & 63.60 & 21.98 & 86.17 & 86.15 
    & 46.20 & 3.94 & 84.78 & 40.83 & 59.24 & 39.56 \\    
    \hline

    GPT-4.1 
    & 5.48 & 1.38 & 27.71 & 8.63 & 14.79 & 10.99 
    & 2.56 & 0.80 & 30.28 & 20.37 & 5.56 & 3.98 \\

    \quad +VerifyMAS    
    &32.39 & 13.29 & 63.92 & 22.25 & 71.32 & 70.71
    & 21.20 & 3.79 & 71.74 & 39.41 & 36.96 & 44.88 \\
    \hline

    Gemini-2.5-Flash 
    & 5.1 & 2.16 & 30.93 & 11.77 & 17.43 & 14.95 
    & 6.13 & 2.42 & 41.48 & 26.23 & 8.55 & 5.83 \\

    \quad +VerifyMAS   
    & 38.48 & 14.43 & 65.23 & 22.56 & 80.25 & 80.62 
    & 36.96 & 4.68 & 80.98 & 39.86 & 64.13 & 53.41 \\
    \hline

    Gemini-2.5-Pro 
    & 4.82 & 2.17 & 29.73 & 11.65 & 14.90 & 12.76 
    & 5.23 & 1.86 & 38.51 & 24.15 & 7.82 & 5.58 \\

    \quad +VerifyMAS  
    & 31.90 & 13.58 & 63.53 & 22.93 & 84.71 & 85.21 
    & 37.50 & 5.33 & 81.52 & 44.63 & 67.39 & 44.24 \\

    \bottomrule
    \end{tabular}
    }
\end{table*}

\subsection{Fine-grained Failure Analysis} \label{app:failure_bar}
\begin{figure}
    \centering
    \includegraphics[width=\linewidth]{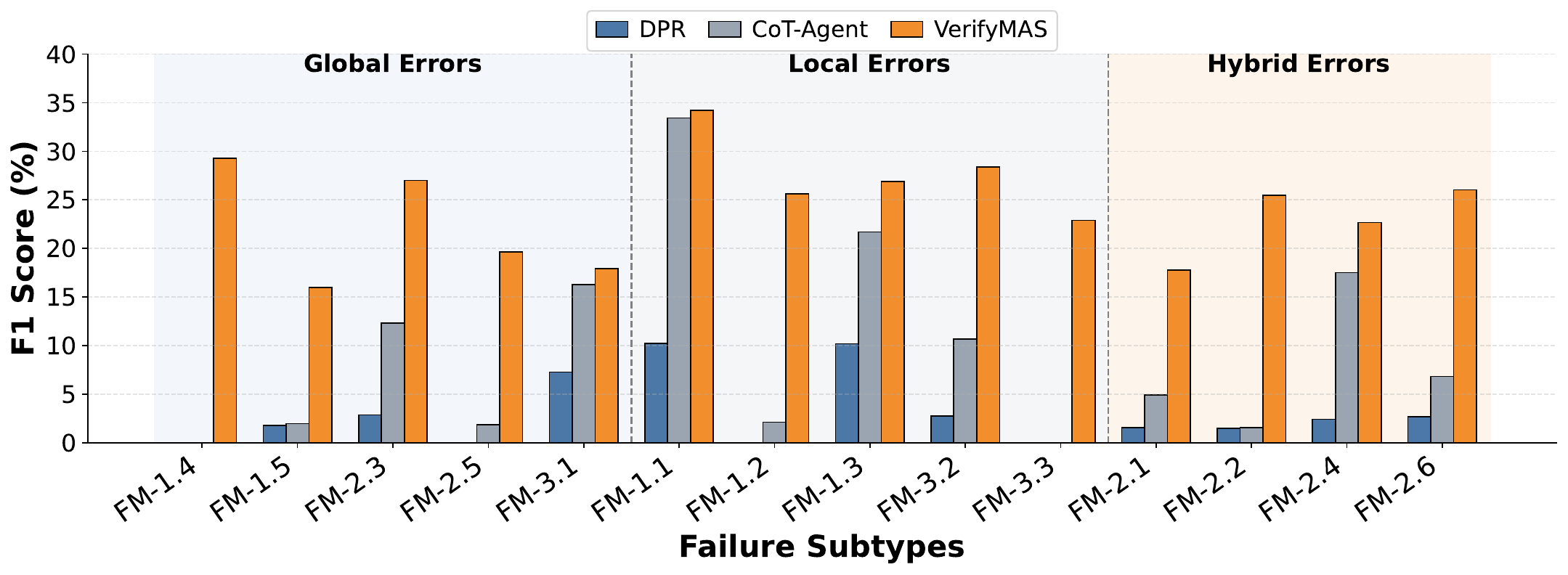}
    \caption{Fine-grained failure analysis grouped by the global, local, and hybrid errors on Aegis-Bench.
    }

    \label{fig:failure_bar2}
    \vspace{-1em}
\end{figure}

\begin{figure}
    \centering
    \includegraphics[width=\linewidth]{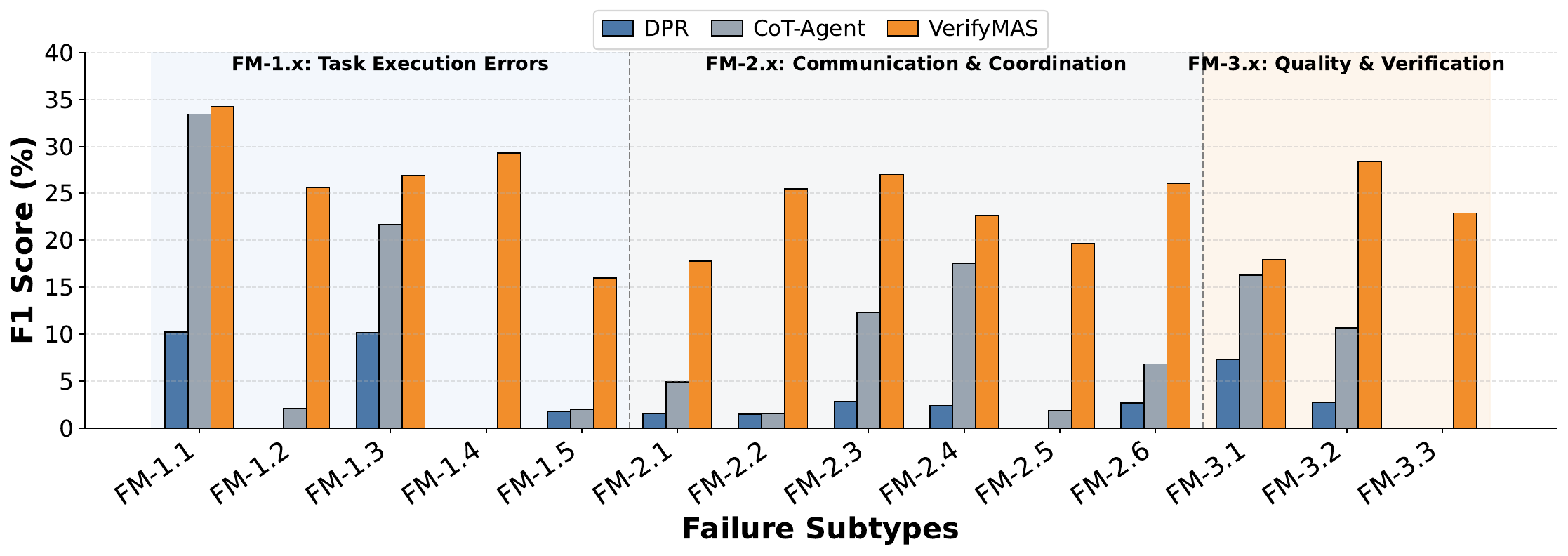}
    \caption{Fine-grained failure analysis grouped by task execution, communication\& Coordination and quality \& verification on Aegis-Bench.
    }
    \label{fig:failure_bar3}
    \vspace{-1em}
\end{figure}

In this section, we report the error attribution performance only (Error-F1), without considering whether the faulty agent is correctly identified in Fig. \ref{fig:failure_bar2} and \ref{fig:failure_bar3}. We observe that model performance varies significantly across error types. Errors related to clear task violations or explicit contradictions (e.g., role deviation or incorrect verification) are generally easier to detect 
while more subtle categories, such as incomplete reasoning, missing coordination, or implicit context neglect, remain challenging across all models. Overall VerifyMAS brings clear gains on three types of errors, indicating that the method improves not only trajectory-level reasoning but also overall robustness in fine-grained failure attribution.

\section{Experiment Details }

\subsection{Training Data Construction Details}  \label{app:failure_mapping}
We construct the training data by converting each annotated trajectory into a set of hypothesis-verification instances. 
\begin{table*}[t]
\centering
\small
\setlength{\tabcolsep}{4pt}
\renewcommand{\arraystretch}{1.15}
\caption{Counter-evidence keyword mapping for constructing contradicted examples. 
For each error type, we list representative lexical cues whose presence indicates that the trajectory is more likely to refute the corresponding error hypothesis.}
\label{tab:contradict_keywords}
\begin{tabularx}{\textwidth}{lX}
\toprule
\textbf{Error Type} & \textbf{Counter-evidence Keywords} \\
\midrule

FM-1.1 
& \texttt{as requested}; \texttt{according to the requirement}; \texttt{followed the requirement}; \texttt{kept the required format}; \texttt{in python}; \texttt{met the specification} \\

FM-1.2 
& \texttt{according to assigned role}; \texttt{within role}; \texttt{as assigned}; \texttt{performed its role}; \texttt{stayed in its role} \\

FM-1.3 
& \texttt{without duplication}; \texttt{without redundant}; \texttt{concise}; \texttt{single final answer}; \texttt{did not repeat} \\

FM-1.4 
& \texttt{as mentioned earlier}; \texttt{based on previous}; \texttt{taking previous context into account}; \texttt{after the correction}; \texttt{considering earlier feedback}; \texttt{from previous turns} \\

FM-1.5 
& \texttt{base case}; \texttt{termination condition}; \texttt{stop when}; \texttt{stopping criterion}; \texttt{return when}; \texttt{until condition is met} \\

FM-2.1 
& \texttt{already completed so skip}; \texttt{no need to repeat}; \texttt{continue from previous result}; \texttt{reuse the previous result}; \texttt{avoid repeating} \\

FM-2.2 
& \texttt{clear instruction}; \texttt{explicit instruction}; \texttt{specifically asked}; \texttt{step-by-step instruction}; \texttt{clearly stated}; \texttt{explicitly stated} \\

FM-2.3 
& \texttt{focus on the main task}; \texttt{directly answer the query}; \texttt{stay on task}; \texttt{relevant to the task}; \texttt{aligned with the goal} \\

FM-2.4 
& \texttt{shared the important information}; \texttt{provided all relevant information}; \texttt{disclosed}; \texttt{made it clear}; \texttt{explicitly mentioned}; \texttt{fully informed} \\

FM-2.5 
& \texttt{incorporated feedback}; \texttt{addressed the concern}; \texttt{responded to the correction}; \texttt{followed the suggestion}; \texttt{took the feedback into account} \\

FM-2.6 
& \texttt{consistent with earlier}; \texttt{same conclusion as before}; \texttt{as stated earlier}; \texttt{remained consistent} \\

FM-3.1 
& \texttt{all requirements are satisfied}; \texttt{after verification}; \texttt{completed all requested parts}; \texttt{finished after checking}; \texttt{fully completed} \\

FM-3.2 
& \texttt{verified}; \texttt{validated}; \texttt{tested}; \texttt{checked}; \texttt{unit test}; \texttt{test case}; \texttt{cross-check} \\

FM-3.3 
& \texttt{verified against expected output}; \texttt{validated with test cases}; \texttt{cross-checked results}; \texttt{checked correctness}; \texttt{confirmed the result} \\

\bottomrule
\end{tabularx}
\end{table*}
\begin{table*}[t]
\centering
\small
\caption{
Nearby error-type mapping for negative hypothesis construction. 
For each annotated error type, nearby error types denote semantically related or easily confusable absent errors. 
}
\label{tab:failure_type_mapping}
\renewcommand{\arraystretch}{1.12}

\begin{tabularx}{0.86\linewidth}{@{}
>{\centering\arraybackslash}p{0.22\linewidth}
>{\centering\arraybackslash}X
@{}}
\toprule
\textbf{Error Type} & \textbf{Nearby Error Types} \\
\midrule
FM-1.1 & FM-2.3, FM-1.4, FM-1.2 \\
FM-1.2 & FM-1.1, FM-2.3, FM-2.5 \\
FM-1.3 & FM-2.1, FM-3.2, FM-2.3 \\
FM-1.4 & FM-2.5, FM-2.6, FM-1.1 \\
FM-1.5 & FM-3.1, FM-3.2, FM-3.3 \\
FM-2.1 & FM-1.3, FM-2.3, FM-2.5 \\
FM-2.2 & FM-2.4, FM-2.3, FM-2.5 \\
FM-2.3 & FM-1.1, FM-2.2, FM-2.1 \\
FM-2.4 & FM-2.2, FM-2.5, FM-1.4 \\
FM-2.5 & FM-1.4, FM-2.4, FM-2.6 \\
FM-2.6 & FM-1.4, FM-2.5, FM-3.3 \\
FM-3.1 & FM-1.5, FM-3.2, FM-3.3 \\
FM-3.2 & FM-3.3, FM-1.5, FM-3.1 \\
FM-3.3 & FM-3.2, FM-2.6, FM-3.1 \\
\bottomrule
\end{tabularx}
\end{table*}
\begin{table*}[t]
\centering
\caption{Distribution of three types of samples in the SFT training set.}
\label{tab:sft_label_distribution}
\begin{tabular}{lccc}
\toprule
\textbf{Label} & \textbf{Type} & \textbf{\# Samples} & \textbf{Ratio} \\
\midrule
A & Entail & 17,753  &  53.02\% \\
B & Neutral & 5,767 & 17.23\% \\
C & Contradict & 9,961  &29.75\% \\
\midrule
\textbf{Total} & -- & \textbf{33,481} & \textbf{100.0\%} \\
\bottomrule
\end{tabular}
\end{table*}

 \begin{table*}[h]
  \centering
  \caption{Training Hyperparameters for HSFT of VerifyMAS }
  \label{tab:training_params}
  \begin{tabular}{lc}
  \toprule
  \textbf{Parameter} & \textbf{SFT Stage} \\
  \midrule
  Training Strategy & FSDP \\
  Learning Rate & 1e-4 \\
  Batch Size (Train) & 64  \\
  Micro Batch Size per GPU & 2  \\
  Max Sequence Length & 8192  \\
  Total Epochs & 2 \\
  LoRA Configuration & $r=64$, $\alpha=16$  \\
  Target Modules & all-linear \\
  Gradient Clipping & 1.0 \\
  Weight Decay & 0.01  \\
  LR Scheduler & cosine  \\
  Warmup Steps Ratio & 0.1  \\
  Model Precision & bf16  \\
  Gradient Checkpointing & True \\
  CPU Offload & True (params)  \\
  Number of GPUs & 2  \\
  \bottomrule
  \end{tabular}
  \end{table*}
\noindent \textbf{Entail Sample Construction.} For each trajectory,  we construct positive entailment samples directly from the ground-truth failure annotations. Specifically, we first extract all annotated faulty agent--error pairs and group them by error type, since the same error hypothesis may be supported by multiple responsible agents. For each ground-truth error type, we instantiate a natural-language error hypothesis using the corresponding template from the predefined error dictionary. The resulting trajectory--hypothesis pair is labeled as entail, indicating that the trajectory provides evidence that this failure occurred. The target supervision contains all gold agents associated with this error type, enabling the model to learn both trajectory-level hypothesis verification and fine-grained multi-agent localization under the same positive hypothesis.

For example, if a given trajectory supports an error hypothesis of error type FM-3.2 and both the Planner and Solver are responsible for this failure, the target output is formatted as:
\[
\bf y =
\begin{aligned}
&\{\texttt{"label"}:\texttt{"A"},\texttt{"agents"}:[\texttt{"Planner"}]\}\\
&\{\texttt{"label"}:\texttt{"A"},\texttt{"agents"}:[\texttt{"Solver"}]\}.
\end{aligned}
\]
For non-entailment samples, the agent set is empty, e.g.,
\[
\{\texttt{"label"}:\texttt{"B"},\texttt{"agents"}:[]\},
\quad
\{\texttt{"label"}:\texttt{"C"},\texttt{"agents"}:[]\}.
\]

\noindent \textbf{Heuristic Search for Negative Sampling.} 
We use two complementary heuristics to construct negative hypotheses. 
First, counter-evidence heuristics search the trajectory for lexical and stage-level cues that explicitly refute an absent error hypothesis. 
Only absent hypotheses with sufficiently strong counter-evidence are labeled as \textit{contradict}.   The detailed counter-evidence for contradiction can be found in the Table \ref{tab:contradict_keywords}. 

Second, we define a nearby failure-type mapping to identify semantically related or easily confusable error types, as shown in Table~\ref{tab:failure_type_mapping}. 
When explicit counter-evidence exists in the trajectory, nearby or absent failure hypotheses are prioritized as contradicted examples; otherwise, nearby hypotheses that are unsupported but not explicitly refuted are used as hard neutral negatives.

\noindent \textbf{Oversampling for Rare Agents.}  During data construction, we also observe a long-tailed distribution over faulty agents. Some agents appear frequently in the training set, while others are associated with only a small number of positive attribution examples. Directly training on this imbalanced data may bias
the model toward frequent agents and weaken its ability to identify rare but valid faulty
agents. To alleviate this issue, we apply oversampling to rare-agent positive instances.
Specifically, for each faulty agent, we count the number of entailment instances in which
the agent is annotated as responsible. Agents with frequencies below a predefined threshold
are treated as rare agents, and their corresponding positive instances are duplicated during
training. This strategy increases the exposure of rare-agent attribution cases without
changing the original label semantics. The ratio of each class for the constructed training dataset are shown in the Table. \ref{tab:sft_label_distribution}.  

\subsection{Training Details} \label{app:training_details}

The details of the hyperparameters are shown in the Table \ref{tab:training_params}. We use the same training hyperparameters for fine-tuning both VerifyMAS-7B and VerifyMAS-8B models.

\section{Algorithm} 
The algorithms of zero-shot failure attribution and hypothesis verification-guided supervised fine-tuning are summarized in Algorithm 1 and Algorithm 2.

\begin{algorithm}[t]
\caption{Zero-shot Failure Attribution}
\label{alg:zeroshot}
\begin{algorithmic}[1]
\REQUIRE Trajectory $\mathcal{\tau}_i$, error type set $\mathcal{Y}$, candidate agent set $\mathcal{A}_i$, LLM verifier $f_{\theta}$
\ENSURE entail error list $\mathcal{Y}_i^{*}$ and selected responsible agents $\mathcal{A}_i^{*}$

\STATE $\mathcal{Y}_i^{*} \leftarrow \emptyset$, $\mathcal{A}_i^{*} \leftarrow \emptyset$

\FOR{each error type $y_m \in \mathcal{Y}$}
    \STATE Construct hypothesis $h_m$ from $y_m$
    \STATE Query $f_{\theta}$ with $(\tau_i, h_m, \mathcal{A}_i, P)$
    \STATE Obtain label $z_{i,m} \in \{\textsc{Entail}, \textsc{Neutral}, \textsc{Contradict}\}$

    \IF{$z_{i,m} = \textsc{Entail}$}
    \STATE $\mathcal{Y}^{*} \leftarrow \mathcal{Y}^{*} \cup \{y_m\}$
    \STATE Obtain responsible agents $\hat{\mathcal{A}}_{i,m} \subseteq \mathcal{A}$
    \STATE $\mathcal{A}_i^{*} \leftarrow \mathcal{A}_i^{*} \cup \{(y_m, a) \mid a \in \hat{\mathcal{A}}_{i,m}\}$
\ENDIF
\ENDFOR

\RETURN $\mathcal{Y}_i^{*}, \mathcal{A}_i^{*}$
\end{algorithmic}
\end{algorithm}

\section{Evaluation Prompts.} \label{app:prompts}

Here we list the prompts used when evaluating various open-source and closed-source models (divided into standard and CoT types). Since Aegis-Bench and Who\&When have unified the data format, the same set of prompts is used for the two benches to ensure the fairness of the assessment. For completeness, we provide the prompts used for the baselines and ablation studies in the supplementary material.

\begin{tcolorbox}[
    title=Error Hypotheses,
    colback=gray!3,
    colframe=black!70,
    fonttitle=\bfseries,
    boxrule=0.6pt,
    arc=2mm,
    left=1mm,
    right=1mm,
    top=1mm,
    bottom=1mm
]

\centering
\small
\renewcommand{\arraystretch}{1.15}

\begin{tabularx}{\linewidth}{>{\bfseries}c X}
\toprule
\textbf{Type} & \textbf{Error Hypothesis} \\
\midrule
FM-1.1 & An agent deviated from the specified task requirements in the trajectory. \\
FM-1.2 & An agent acted outside its designated role in the trajectory. \\
FM-1.3 & An agent introduced redundant or unnecessary steps in the trajectory. \\
FM-1.4 & An agent overlooked important context from earlier conversation history in the trajectory. \\
FM-1.5 & An agent failed to maintain a proper termination condition in the trajectory. \\
FM-2.1 & An agent repeated a task that had already been handled in the trajectory. \\
FM-2.2 & An agent made a request unclear or ambiguous for other agents in the trajectory. \\
FM-2.3 & An agent drifted away from the main goal of the task in the trajectory. \\
FM-2.4 & An agent omitted important information needed by other agents in the trajectory. \\
FM-2.5 & An agent overlooked input or corrections from other agents in the trajectory. \\
FM-2.6 & An agent used reasoning that was inconsistent with earlier statements in the trajectory. \\
FM-3.1 & An agent ended the task before all requirements were met in the trajectory. \\
FM-3.2 & An agent skipped necessary verification or validation steps in the trajectory. \\
FM-3.3 & An agent performed flawed or unreliable verification in the trajectory. \\
\bottomrule
\end{tabularx}

\end{tcolorbox}

\clearpage
\begin{tcolorbox}[
    title=Evaluation and SFT Prompt-VerifyMAS,
    colback=white,
    colframe=black!65,
    coltitle=white,
    colbacktitle=black!70,
    fonttitle=\bfseries,
    arc=2mm,
    boxrule=0.8pt,
    left=2mm,
    right=2mm,
    top=2mm,
    bottom=2mm
]

You are a careful verifier for multi-agent trajectory failure attribution. You will be given: A trajectory; A failure hypothesis; The list of candidate agents appearing in the trajectory. Your task is to jointly predict:
\begin{itemize}
    \item \textbf{label}: one of A, B, C
    \item \textbf{agents}: responsible agent(s) for the hypothesized failure
\end{itemize}
\textbf{Label meanings:}
\textbf{A} = entail, \textbf{B} = neutral, \textbf{C} = contradict

\textbf{Decision rules:}
\begin{itemize}
    \item Choose \textbf{A} when the trajectory provides clear or reasonably strong evidence that the hypothesized failure occurred, and that it negatively affected, or likely affected, the final outcome, final answer, final decision, or successful task completion.
    
    \item Choose \textbf{B} when the evidence is mixed, incomplete, weak, or the impact on the final outcome is uncertain.
    
    \item Choose \textbf{C} when the trajectory provides clear evidence that the hypothesis is false or inconsistent with what actually happened.
    
    \item If the failure appears to have been minor, corrected later, or not connected to the final outcome, prefer \textbf{B} over \textbf{A}.
    
    \item Use \textbf{C} only when the hypothesis is clearly contradicted by the trajectory, not merely because support is weak.
\end{itemize}

\textbf{Agent attribution rules:}
\begin{itemize}
    \item A failure may be caused by one agent or multiple agents.
    
    \item If label = \textbf{A}, output one JSON object for each likely responsible agent.
    
    \item Each JSON object should contain exactly one responsible agent in the ``agents'' list.
    
    \item Include an agent only when there is clear or reasonably strong evidence that this agent contributed to the hypothesized failure.
    
    \item Do not include agents merely because they are mentioned near the error.
    
    \item Do not include agents that only followed instructions from another faulty agent unless their own action also contributed to the failure.
    
    \item Only output agent names that appear exactly in the provided candidate agent list.
    
    \item If label = \textbf{A} but no responsible agent can be confidently identified, output one JSON object with ``agents'': [].
    
    \item If label is \textbf{B} or \textbf{C}, output exactly one JSON object with ``agents'': [].
\end{itemize}

\textbf{Output format:}

Do not output thinking, reasoning, analysis, or \texttt{<think>...</think>}. Return JSON objects only. If there are multiple responsible agents, return multiple JSON objects, one per line.

Each JSON object must have exactly these keys:

\begin{verbatim}
{"label":"A","agents":["agent_name"]}
\end{verbatim}

\textbf{Valid examples:}

\begin{verbatim}
{"label":"A","agents":["Planner"]}
{"label":"A","agents":["Solver"]}
{"label":"B","agents":[]}
{"label":"C","agents":[]}
\end{verbatim}

Do not output explanations.

Do not output markdown.

Do not output extra text.

\end{tcolorbox}

\begin{algorithm}[tb]
\caption{Hypothesis Verification-guided Supervised Fine-tuning}
\label{alg:sft}
\begin{flushleft}
\hspace*{\algorithmicindent}\textbf{Input}: Training trajectories $\mathcal{D}=\{\tau_i\}_{i=1}^{N}$, error taxonomy $\mathcal{Y}$, base LLM $f_{\theta}$

\hspace*{\algorithmicindent}\textbf{Output}: Fine-tuned model $f_{\theta^\ast}$
\end{flushleft}

\begin{algorithmic}[1]

\STATE Initialize SFT dataset $\mathcal{D}_{\mathrm{sft}} \leftarrow \emptyset$

\FOR{each trajectory $\tau_i \in \mathcal{D}$}
    \STATE Extract candidate agents $\mathcal{A}_i$ from trajectory $\tau_i$
    \STATE Let $\mathcal{G}_i=\{(e,a)\}$ denote the ground-truth faulty agent-error pairs

    \FOR{each ground-truth pair $(e, a) \in \mathcal{G}_i$}
        \STATE Construct an entailment hypothesis $h_i^{A}$ for error type $e$
        \STATE Set target response $y_i^{A}=\{\texttt{"label"}:\texttt{"A"},\texttt{"agent"}:[a]\}$
        \STATE Add $(\tau_i,h_i^{A}, \mathcal{A}_i, P, y_i^{A})$ to $\mathcal{D}_{\mathrm{sft}}$
    \ENDFOR

    \STATE Sample contradiction error types $\mathcal{Y}_i^{C} \subseteq \mathcal{Y} \setminus \{e \mid (e, a)\in\mathcal{G}_i\}$

 \STATE Construct contradicted error types $\mathcal{Y}_i^{C}$ using explicit counter-evidence and the nearby failure-type mapping $\mathcal{M}_{\mathrm{near}}$

\FOR{each error type $y^{C} \in \mathcal{Y}_i^{C}$}
    \STATE Construct a contradiction hypothesis $h_i^{C}$ for $y^{C}$
    \STATE Set target response $y_i^{C}=\{\texttt{"label"}:\texttt{"C"},\texttt{"agents"}:[]\}$
    \STATE Add $(\tau_i,h_i^{C},\mathcal{A}_i, P,y_i^{C})$ to $\mathcal{D}_{\mathrm{sft}}$
\ENDFOR

\STATE Sample neutral error types $\mathcal{Y}_i^{B}$ from nearby but unsupported types in $\mathcal{M}_{\mathrm{near}}$

\FOR{each error type $y^{B} \in \mathcal{Y}_i^{B}$}
    \STATE Construct a neutral hypothesis $h_i^{B}$ for $y^{B}$
    \STATE Set target response $y_i^{B}=\{\texttt{"label"}:\texttt{"B"},\texttt{"agents"}:[]\}$
    \STATE Add $(\tau_i,h_i^{B},\mathcal{A}_i, P,y_i^{B})$ to $\mathcal{D}_{\mathrm{sft}}$
\ENDFOR
\ENDFOR

\FOR{each training step}
    \STATE Sample a mini-batch $\mathcal{B}$ from $\mathcal{D}_{\mathrm{sft}}$
    \STATE Format each instance as input ${{\bf{x}}_{i,m}}= \{\tau_i, h_m, P, \mathcal{A}_i\}$ with  trajectory corresponding to failure hypothesis $h_m$  and target response $\textbf{y}_{i,m}$
    \STATE Compute LM loss $\mathcal{L}_{\mathrm{HSFT}}=-\frac{1}{|\mathcal{B}|}\sum_{(\textbf{x}_{i,m},\textbf{y}_{i,m})\in\mathcal{B}}\sum_{t=1}^{|\mathbf{y}_{i,m}^*|}
\log p_{\theta}
\left(
y_{i,m}^{(t)*}
\mid
{\bf x}_{i,m}, y_{i,m}^{(<t)*}
\right)$


    \STATE Update model parameters $\theta$ by minimizing $\mathcal{L}_{\mathrm{HSFT}}$
\ENDFOR

\RETURN Fine-tuned model $f_{\theta^\ast}$

\end{algorithmic}
\end{algorithm}

\end{document}